%% file: main.tex
\documentclass{jingdong}
\usepackage{amsmath,amssymb,graphicx,booktabs}
\usepackage{xcolor,colortbl}
\colorlet{comavrow}{jingdongred!7!white}
\colorlet{tablesection}{jingdongred!4!white}
\newcommand{\paperth}[1]{\textbf{#1}}
\definecolor{paperrowgray}{RGB}{210,210,210}
\newcommand{\paperrowsep}{\arrayrulecolor{paperrowgray}\specialrule{0.22pt}{0.4pt}{0.4pt}\arrayrulecolor{black}}
\usepackage{url}
\usepackage{placeins}
\usepackage{afterpage}
\newcommand{\cth}[1]{\multicolumn{1}{c}{#1}}

\title{HumanCtrl-Bench: Evaluating Controllability in Human-Centric Audiovisual Generation and Editing}
\author[*]{Xiyang Li}
\author[*]{Tengjie Li}
\author[*]{Tao Liu}
\author{Rong Qin}
\author[\dagger]{Ke Lin}
\author{Youzheng Wu}
\author{Xiaodong~He}
\affiliation{JD Technology, JD}
\date{26 November 2026}

\abstract{Human-centered content production needs to reuse recognizable characters and voices while changing dialogue, actions, and scenes. This requires models to selectively control appearance, performance, and speech, bind changes to the intended actors, and preserve specified source content. Existing benchmarks do not comprehensively assess human controllability across reference-conditioned generation and source-video editing, particularly attribute transfer, actor binding, and source preservation. We introduce HumanCtrl-Bench, covering eight generation and editing families with 606 cases: 298 single-subject and 308 multi-subject instances with explicit reference roles, dialogue assignments, and editing scopes. We evaluate 3,030 outputs from five models through task-condition measurements, multimodal large language model (MLLM) evaluations, and quality estimators. Wan 3.0 ranks first by overall mean rank in both task-condition measurements and MiMo-based MLLM assessment on their respective comparable subsets. Kling 4.0 leads several visual quality and audio production quality estimators.}

\newcommand{\benchmarkoverview}{%
  \par\noindent\begin{minipage}{\linewidth}
  \centering
  \captionsetup{font=footnotesize}
  \includegraphics[width=0.84\linewidth]{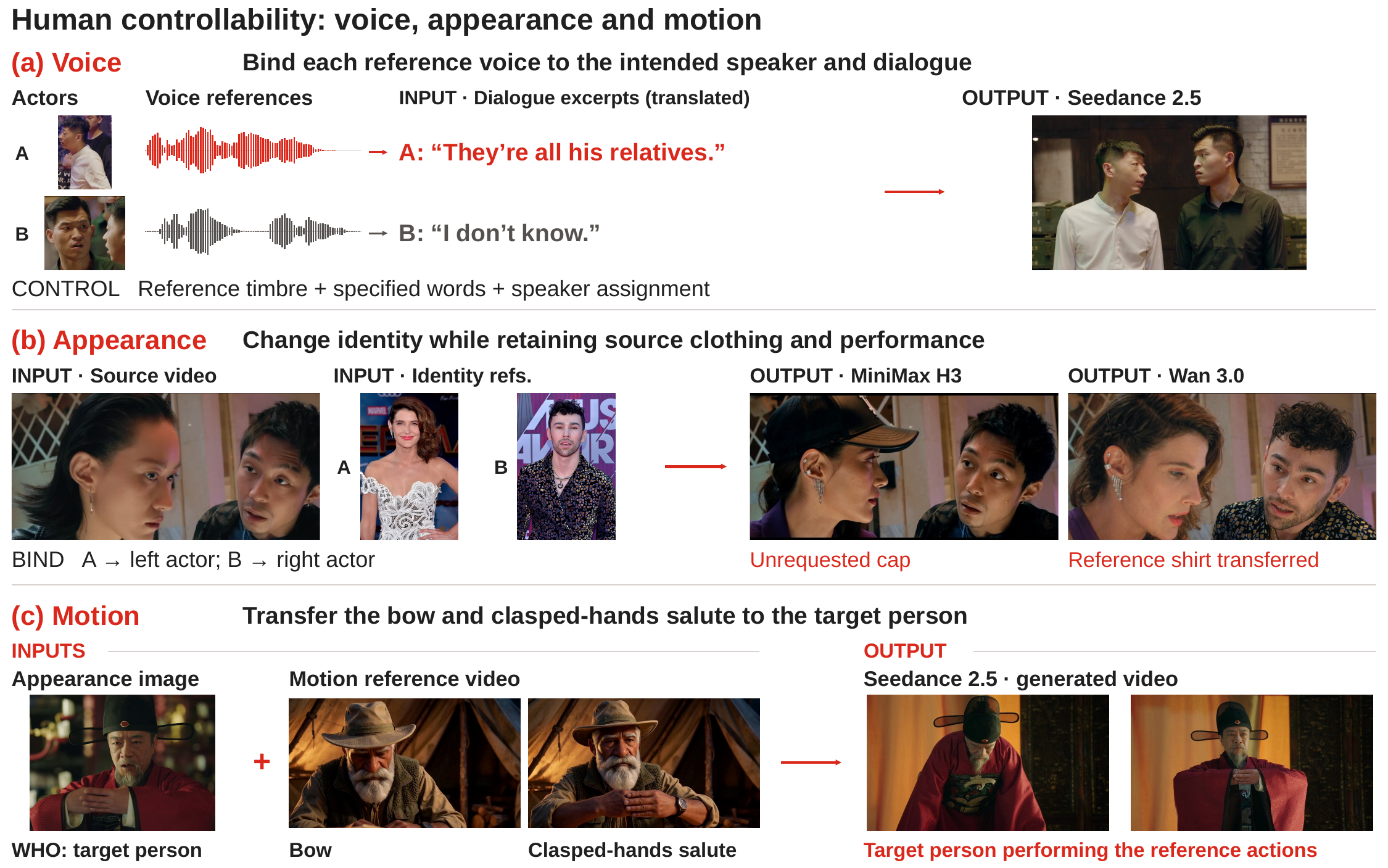}
  \captionof{figure}{Human control along three dimensions in HumanCtrl-Bench. (a) Voice references and target dialogue are assigned to individual speakers; waveforms depict the reference audio, and dialogue excerpts are translated from the Chinese task caption. (b) Identity references specify which person to change while source clothing and performance must remain. The MiniMax H3~\citep{minimax2026h3} output introduces a cap, while Wan 3.0~\citep{alibabacloud2026wan30} transfers a reference shirt. (c) Left: an appearance image specifies the target person, and a separate video specifies the bow and clasped-hands salute. Right: sampled Seedance 2.5~\citep{bytedanceseed2026seedance25} output frames show the target person performing these actions. Arrows distinguish supplied inputs from generated outputs. Voice and identity outputs show frames at 3\,s; the motion sequence shows frames near the beginning and end.}
  \label{fig:benchmark-overview}
  \end{minipage}\par\vspace{4mm}}

\renewcommand{\maketitle}{%
  \begingroup
  \setlength{\parskip}{0pt}
  \centering
  \jingdongred{\hrule height 0.5pt}
  \vspace{2mm}
  \titlelist\par
  \vspace{2mm}
  \jingdongred{\hrule height 0.5pt}
  \vspace{2mm}
  {\bfseries\authorlist\par}
  \vspace{1.5mm}
  \affiliationlist\par
  \vspace{2mm}
  \endgroup
  \begin{tcolorbox}[enhanced,colback=white,colframe=jingdongbg,
    boxrule=0.5pt,arc=2mm,left=3mm,right=3mm,top=2mm,bottom=2mm,
    before skip=0pt,after skip=3mm]
  \setlength{\parskip}{0pt}
  \setlength{\parindent}{0pt}
  {\abstractfont\sffamily\bfseries\jingdongred{Abstract}\par}
  \vspace{1mm}
  \renewcommand{\abstractinfont}{\fontsize{9.5}{11}\selectfont}
  \abstractlist\par
  \vspace{1mm}
  \checkdatalist
  \end{tcolorbox}
  \thispagestyle{firststyle}
}

\begin{document}
\maketitle
\begingroup\renewcommand{\thefootnote}{}\footnotetext{\textsuperscript{*}Equal contribution.\quad\textsuperscript{\ensuremath{\dagger}}Project leader.}\endgroup
\benchmarkoverview
\setlength{\emergencystretch}{2em}
\raggedbottom
\setcounter{topnumber}{3}
\renewcommand{\topfraction}{0.9}
\renewcommand{\textfraction}{0.08}
\makeatletter
\setlength{\@fptop}{0pt}
\setlength{\@fpsep}{12pt}
\setlength{\@fpbot}{0pt plus 1fil}
\makeatother
\setlength{\textfloatsep}{12pt plus 2pt minus 2pt}

\clearpage
\section{Introduction}
\label{sec:introduction}
\afterpage{\begin{table}[!t]
\caption{Coverage of reference control and source-video editing in related benchmarks. Check marks denote settings explicitly evaluated in the benchmark documentation.}
\label{tab:positioning}
\centering

\footnotesize
\setlength{\tabcolsep}{2.9pt}
\renewcommand{\arraystretch}{1.0}
\begin{tabular*}{\linewidth}{@{\extracolsep{\fill}}l*{9}{c}@{}}
\arrayrulecolor{jingdongred}\toprule\arrayrulecolor{black}

& \multicolumn{2}{c}{Appearance refs.}
& \multicolumn{2}{c}{Voice refs.}
& \multicolumn{2}{c}{Performance}
 & \multicolumn{3}{c}{Source-video editing} \\
\cmidrule(lr){2-3}\cmidrule(lr){4-5}\cmidrule(lr){6-7}\cmidrule(lr){8-10}
\cth{Benchmark} & \cth{Single} & \cth{Multi} & \cth{Single} & \cth{Multi} & \cth{Transfer} & \cth{+ Voice} & \cth{Visual} & \cth{Audio} & \cth{Retention} \\
\midrule
OpenS2V-Eval~\citep{yuan2025opens2vnexus} & $\checkmark$ & $\checkmark$ & & & & & & & \\
Phantom test set~\citep{Liu_2025_ICCV} & $\checkmark$ & $\checkmark$ & & & & & & & \\
MSAVBench~\citep{wei2026msavbench} & $\checkmark$ & & $\checkmark$ & & & & & & \\
MultiRef-Compass~\citep{zhang2026multirefcompass} & $\checkmark$ & $\checkmark$ & $\checkmark$ & & & & & & \\
ORAV~\citep{hua2026orav} & $\checkmark$ & $\checkmark$ & $\checkmark$ & & $\checkmark$ & $\checkmark$ & & & \\
OmniVBench~\citep{li2026omnivbench} & $\checkmark$ & $\checkmark$ & & & $\checkmark$ & & & & \\
VACE-Benchmark~\citep{jiang2025vaceallinonevideocreation} & $\checkmark$ & & & & $\checkmark$ & & $\checkmark$ & & $\checkmark$ \\
CoVEBench~\citep{wu2026covebenchvideoeditingmodels} & & & & & & & $\checkmark$ & & $\checkmark$ \\
OmniEdit-Bench~\citep{miao2026omnieditbenchcomprehensivebenchmarkinstructionbased} & & & & & & & $\checkmark$ & $\checkmark$ & $\checkmark$ \\
\midrule
\textbf{HumanCtrl-Bench} & $\checkmark$ & $\checkmark$ & $\checkmark$ & $\checkmark$ & $\checkmark$ & $\checkmark$ & $\checkmark$ & $\checkmark$ & $\checkmark$ \\
\bottomrule
\end{tabular*}
\end{table}}

E-commerce presentation, game-character production, and virtual companionship require audiovisual content centered on recurring people~\citep{kong2025multitalk,li2025hunyuangameindustrialgradeintelligentgame,xu2024vasa1lifelikeaudiodriventalking}. Characters must adapt to new dialogue, actions, and scenes while retaining recognizable identities and voices. Editing an existing clip also calls for reusing its performance and other specified content. Models therefore need to selectively change human attributes while maintaining the remaining attributes and their assignment across people. Recent audiovisual generators accept combinations of images, performance videos, voice samples, and text~\citep{seedance2026seedance20advancingvideo,klingteam2025klingomnitechnicalreport,li2026mmcontrol}, and video-editing interfaces extend these controls to existing performances~\citep{jiang2025vaceallinonevideocreation,luma2025ray3modify}. Evaluation must establish whether these controls act on the intended people.

Human control is challenging because each reference usually contains several co-occurring attributes. A performance video supplies actions alongside the demonstrator's identity and setting; a voice sample supplies timbre together with words that may differ from the requested dialogue~\citep{hua2026orav,li2026omnivbench,li2026mmcontrol}. The model must extract the required information, such as reproducing a demonstrated action with a target person or speaking new words in a reference voice. Multi-person scenes further require these attributes to remain assigned to the correct actors through movement and alternating speaking turns~\citep{kong2025multitalk,zhou2026mtavgbench,wei2026msavbench}. Reference resemblance must therefore be considered together with the intended actor assignment.

Editing makes the scope of control explicit: specified content must survive the modification~\citep{polyak2025moviegencastmedia}. Identity replacement may change the face, hair, and visible build while retaining clothing, performance, and soundtrack. Background replacement may adapt a new environment while retaining foreground people and their object interactions. Figure~\ref{fig:benchmark-overview} organizes human control along three dimensions: voice, appearance, and motion. Voice references are bound to individual speakers and target dialogue; identity references modify selected appearance attributes while preserving source garments; motion references guide the performance of a separately specified person. The identity-editing examples reveal unrequested accessories and reference-clothing transfer. Across these dimensions, task definitions connect three requirements: {\fontspec{LiberationSerif-BoldItalic.ttf}[Path=fonts/] attribute transfer} specifies what each reference contributes, {\fontspec{LiberationSerif-BoldItalic.ttf}[Path=fonts/] actor binding} specifies to whom it belongs, and {\fontspec{LiberationSerif-BoldItalic.ttf}[Path=fonts/] source preservation} specifies what editing must retain.

Existing benchmarks advance different parts of this evaluation. Reference-conditioned suites, including ORAV~\citep{hua2026orav} and MultiRef-Compass~\citep{zhang2026multirefcompass}, examine identity fidelity, multi-entity binding, and combinations of appearance, performance, and voice~\citep{yuan2025opens2vnexus,wei2026msavbench,li2026omnivbench}. Editing suites, including VACE-Benchmark~\citep{jiang2025vaceallinonevideocreation}, CoVEBench~\citep{wu2026covebenchvideoeditingmodels}, and OmniEdit-Bench~\citep{miao2026omnieditbenchcomprehensivebenchmarkinstructionbased}, evaluate requested modifications, source retention, and audio editing. These benchmarks do not yet comprehensively cover human controllability across reference-conditioned generation and source-video editing. Table~\ref{tab:positioning} identifies the intersection targeted here: a common task suite covering single- and multi-person appearance and timbre control together with selective source-video modification and preservation. This organization supports comparisons across generation and editing under shared human-control requirements.

We introduce \textbf{HumanCtrl-Bench}, comprising 606 instances across eight families, with 298 single-subject and 308 multi-subject cases. Four generation families combine appearance, performance, and voice conditions; four editing families change appearance, voice, articulation, or background within a defined scope. Each instance records reference roles, target actors, dialogue assignments, and protected attributes. Model-adapted captions, input mappings, and submitted requests support tracing task definitions to generation records. Figure~\ref{fig:task-overview} illustrates single- and multi-person examples from all eight families.

We evaluate five models on 3,030 generated outputs for the same 606 cases using task-condition measurements, multimodal large language model (MLLM) evaluations, and objective quality estimators. Condition measurements select applicable metrics from reference roles and editing scopes, reporting native scales and metric-specific common-case coverage by task and subject count. MLLMs supply independent task-conditioned checklist judgments, and quality estimators characterize appearance, motion, and audio production. These views examine requested conditions, perceived quality, and automatic-evaluator behavior separately.

Wan 3.0~\citep{alibabacloud2026wan30} ranks first by overall mean rank in both task-condition measurements and MiMo-based MLLM assessment on their respective comparable subsets (Tables~\ref{tab:condition-main} and~\ref{tab:mimo-checklist}). Kling 4.0~\citep{kuaishou2026kling40} leads several visual quality and audio production quality estimators (Table~\ref{tab:quality-diagnostics}). Selected edits reveal omitted modifications and unintended clothing changes (Figure~\ref{fig:preservation-failures}).

Our contributions are:
\begin{itemize}
\item We introduce HumanCtrl-Bench for human controllability, covering single- and multi-person reference-conditioned generation and source-video editing in one suite organized around attribute transfer, actor binding, and source preservation.
\item We develop a task-conditioned evaluation protocol that associates measurements with reference roles, actor assignments, and editing scopes, and combines them with MLLM evaluations, and quality estimators to characterize condition satisfaction and audiovisual quality.
\item We evaluate five audiovisual generators on HumanCtrl-Bench, analyzing capability differences and failure modes across tasks and subject settings. Editing examples show reference-like clothing changes outside the requested scope. Reproducible measurements and coverage records support further identity, voice, and source-preservation diagnostics.
\end{itemize}

\section{Related Work}
\label{sec:related-work}

\textbf{Audiovisual generation and control.} Joint audio-video models synthesize visual content, speech, and environmental sound~\citep{seedance2026seedance20advancingvideo,deepmind2026veo3,hacohen2026ltx2efficientjointaudiovisual,duan2026longhorizonaudiovisualgenerationpersistent}. Multimodal interfaces further expose image, video, and audio references for conditioning or editing~\citep{berniniteam2026berninilatentsemanticplanning,klingteam2025klingomnitechnicalreport,li2026mmcontrol,xai2026imaginevideo15references}. These interfaces motivate evaluating the requested contribution of each input and its assignment to the intended actor.

\textbf{ Avatar Generation.} Avatar generation methods control identity and performance through appearance, motion, and speech conditions. Speech-driven portrait methods animate articulation, expression, and head motion~\citep{prajwal2020lipsyncexpertneed,zhang2023sadtalkerlearningrealistic3d,xu2024vasa1lifelikeaudiodriventalking,tian2024emoemoteportraitalive,xu2024hallohierarchicalaudiodrivenvisual,chen2024echomimiclifelikeaudiodrivenportrait}. Appearance- and motion-conditioned animation separates the identity reference from the driving pose or performance~\citep{hu2024animateanyoneconsistentcontrollable,guo2025liveportraitefficientportraitanimation}. Audio-driven generation further encompasses body and scene motion~\citep{lin2025omnihuman1rethinkingscalinguponestage,gan2025omniavatarefficientaudiodrivenavatar,wang2025fantasytalkingrealistictalkingportrait,gao2025wans2vaudiodrivencinematicvideo,wang2026joystreamerunlockinghighlyexpressive}, with related work addressing multimodal subject control, multi-person conversation, and streaming avatars~\citep{chen2025humohumancentricvideogeneration,li2026mmcontrol,kong2025multitalk,zhu2026omniliveavatarminutelevelrealtimestreaming,li2025joystreamerflashrealtimeinfiniteaudiodriven}. Editing adds the requirement of making selected changes while retaining specified source content. General video-editing methods address temporal consistency, edit propagation, and reference-guided source modification~\citep{geyer2023tokenflowconsistentdiffusionfeatures,yang2023rerendervideozeroshottextguided,ku2024anyv2vtuningfreeframeworkvideotovideo,jiang2025vaceallinonevideocreation,xiao2026joyaivideoeditrealtimeopenendedvideo}; lip-editing methods modify articulation using a supplied recording~\citep{cheng2022videoretalkingaudiobasedlipsynchronization,li2025latentsynctamingaudioconditionedlatent}. These interfaces call for distinguishing timbre references from driving recordings and evaluating requested changes together with source preservation.

\textbf{ Video Generation Benchmarks.} Video generation benchmarks differ in the controls they test and the evidence they use to assess success. General suites evaluate perceptual quality, semantics, compositional relations, and physical behavior~\citep{huang2024vbench,liu2024evalcrafter,zheng2025vbench2,sun2025t2vcompbench,NEURIPS2025_4ec03ed0}. Evaluations of subject-conditioned generation assess reference fidelity and multi-subject control~\citep{yuan2025opens2vnexus,Liu_2025_ICCV,chen2025multisubject}; audiovisual suites examine generation quality, dialogue, sound-source relations, and consistency across shots~\citep{hua2026vabench,zhou2026mtavgbench,liu2026castbench,wei2026msavbench}. Learned assessors provide quality or preference signals~\citep{he2024videoscore,he2025videoscore2,liu2025videoreward,xu2026visionreward,bansal2025videophy2}, while rubric-guided judging and position-bias studies inform automatic-evaluation protocols~\citep{liu2023geval,zheng2023mtbench,wang2024fair}. The closest reference benchmarks assess semantic roles and target assignment, heterogeneous-reference composition, and visual-factor preservation and routing~\citep{hua2026orav,zhang2026multirefcompass,li2026omnivbench}. HumanCtrl-Bench brings these evaluation questions to actor-specific visual and vocal control across generation and selective editing. It includes single- and multi-person appearance and timbre references, together with explicit requirements to retain source clothing, actions, or audio. Table~\ref{tab:positioning} compares the documented task coverage.

\section{HumanCtrl-Bench}
\label{sec:data}
\label{sec:tasks}

\subsection{Task Taxonomy}
\textbf{Condition specification.} Each case describes the output through three requirements. Attribute transfer identifies what to take from each image, video, audio sample, or text description. Actor binding assigns those properties and spoken lines to people in the output. Source preservation identifies the content that must remain during editing. A voice sample can thus specify timbre while the caption supplies new words; a source video can specify performance while an image changes identity.

\begin{figure}[!t]
\centering
\includegraphics[width=0.90\linewidth]{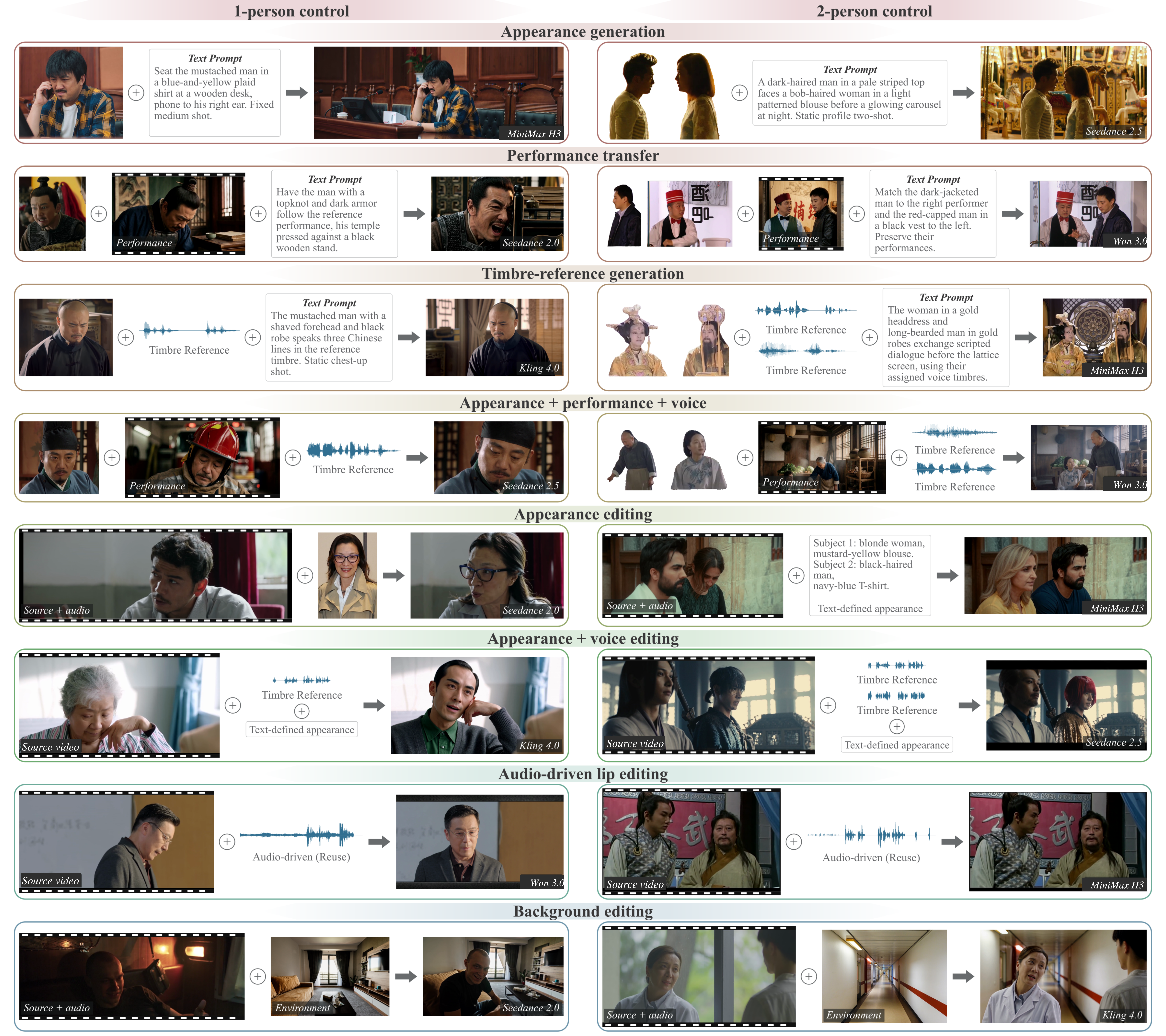}
\caption{The eight task families of HumanCtrl-Bench, illustrated by one-person and two-person cases. Each row connects reference inputs and task instructions to a saved model output. Timbre references specify voice characteristics; Audio-driven (Reuse) supplies a complete recording. Text prompts are condensed for display.}
\label{fig:task-overview}
\end{figure}

\textbf{Task organization.} We group these requirements into four generation and four editing families (Figure~\ref{fig:task-overview}). Distinct appearance-editing scopes and background-guidance modes give eleven fine-grained tasks, each with single- and multi-subject cases (Table~\ref{tab:tasks}). These 22 construction settings define the input conditions. The main results use the eight families split by subject count, yielding 16 reporting groups; person replacement combines its three variants only after applying their respective measurement rules.

\begin{table}[!t]
\caption{Task taxonomy and case counts in HumanCtrl-Bench. The eight families comprise eleven control tasks; Single and Multi indicate the number of controlled or referenced principal actors.}
\label{tab:tasks}
\centering
\small

\setlength{\tabcolsep}{2.3pt}
\renewcommand{\arraystretch}{1.10}
\begin{tabular}{@{}>{\raggedright\arraybackslash}m{0.17\linewidth}>{\raggedright\arraybackslash}m{0.365\linewidth}>{\raggedright\arraybackslash}m{0.22\linewidth}@{\hspace{3pt}}c@{\hspace{7pt}}c@{\hspace{7pt}}c@{}}
\arrayrulecolor{jingdongred}\toprule\arrayrulecolor{black}

\centering\arraybackslash \textbf{ Family} & \centering\arraybackslash \textbf{ Control task} & \centering\arraybackslash \textbf{ Reference inputs} & \cth{\textbf{ Single}} & \cth{\textbf{ Multi}} & \cth{\textbf{ Total}} \\
\midrule
\rowcolor{comavrow}
\rowcolor{tablesection}
\multicolumn{6}{l}{\textbf{ Reference-conditioned generation}\quad 4 families, 4 tasks} \\
\addlinespace[3pt]
\textbf{ Appearance} & Text-directed performance & Images & 58 & 49 & 107 \\
\paperrowsep
\addlinespace[3pt]
\textbf{ Performance} & Transfer performance & Images, \mbox{performance video} & 41 & 52 & 93 \\
\paperrowsep
\addlinespace[3pt]
\textbf{ Voice} & Speak with assigned timbre & Images, \mbox{timbre samples} & 59 & 78 & 137 \\
\paperrowsep
\addlinespace[3pt]
\textbf{ Joint control} & Combine appearance and\newline performance with timbre & Images, \mbox{performance video}, \mbox{timbre samples} & 19 & 18 & 37 \\
\addlinespace[5pt]
\rowcolor{comavrow}
\rowcolor{tablesection}
\multicolumn{6}{l}{\textbf{ Source-video editing}\quad 4 families, 7 tasks} \\
\addlinespace[3pt]
\textbf{ Appearance} & Replace face; retain clothing & \mbox{Source video}, images & 9 & 10 & 19 \\
\paperrowsep
\addlinespace[2pt]
 & Replace identity, hair and \mbox{visible build}; retain clothing & \mbox{Source video}, images & 2 & 2 & 4 \\
\paperrowsep
\addlinespace[2pt]
 & Text edit; retain source audio & \mbox{Source video} & 52 & 40 & 92 \\
\paperrowsep
\addlinespace[4pt]
\textbf{ Appearance\newline and voice} & Edit appearance and voice & \mbox{Source video}, \mbox{timbre samples} & 28 & 32 & 60 \\
\paperrowsep
\addlinespace[4pt]
\textbf{ Lip editing} & Match speech; retain recording & \mbox{Source video}, \mbox{full recording} & 10 & 8 & 18 \\
\paperrowsep
\addlinespace[4pt]
\textbf{ Background} & Adapt to source camera and lighting & \mbox{Source video}, \mbox{background image} & 10 & 10 & 20 \\
\paperrowsep
\addlinespace[2pt]
 & Match structures and layout & \mbox{Source video}, \mbox{background image} & 10 & 9 & 19 \\
\midrule
\textbf{ Total} & \multicolumn{2}{l}{11 tasks, 22 task/subject settings} & \textbf{ 298} & \textbf{ 308} & \textbf{ 606} \\
\bottomrule
\end{tabular}
\par\smallskip
\raggedright\footnotesize Images specify appearance. Video-audio use follows the task-specific input configuration; timbre samples specify voice characteristics, while full recordings specify the output soundtrack.
\end{table}

\subsubsection{Reference-conditioned generation}
\textbf{Appearance generation.} Images define the characters' identities and styling, while text specifies actions and interactions. The task tests whether appearance remains consistent as expression and movement change. In multi-person cases, each appearance must remain associated with its assigned behavior, such as handling an object or responding to another actor.

\textbf{Performance transfer.} Appearance images are paired with a performance-reference video. The model transfers the demonstrated actions, expressions, and their progression to the image-defined characters; captioned dialogue guides speech generation. Multi-person cases bind each target actor to a source performer, preserving who gestures, listens, or interacts with an object as the performance unfolds.

\textbf{Voice-reference generation.} Appearance images and actor-specific timbre samples accompany the requested dialogue. The output must combine the assigned appearance, voice, and spoken words. In multi-person exchanges, these assignments continue across speaking turns. Paired cases use the same reference media and speaker assignments but different target dialogue, testing whether models can preserve the reference voice while speaking new words.

\textbf{Joint reference generation.} Images, a performance-reference video, timbre samples, and dialogue text jointly specify appearance, actions, and voice. The requirements meet in the same actor. In one two-person cooking case, the standing actor continues stirring while answering the seated actor: appearance, ongoing motion, and the active speaking role must remain coordinated.

\subsubsection{Source-video editing}
\textbf{Person replacement.} Image-guided tasks specify the target person through a reference image. Depending on the instruction, the model must replace either the face alone or the face, hairstyle, and visible body build, while preserving the source clothing. Text-guided tasks instead describe the desired appearance in words and define which attributes should change. The source performance and soundtrack are required to remain unchanged. In multi-person scenes, each edit must be applied to the designated person.

\textbf{Joint appearance and voice editing.} A source video supplies performance and scene layout, text specifies new appearances and dialogue, and timbre samples supply the voices. The edited appearances and voices must stay associated with the same source actors. The assignments also specify silence: in one two-person case both actors have timbre samples, but only one is instructed to speak.

\textbf{Supplied-recording lip editing.} A source video and a complete recording define local lip and jaw edits. The recording is retained in full, together with the source identity, clothing, gaze, body motion, and camera movement. Nine early/late pairs vary speech onset while keeping the recorded content and speech rate fixed. Outputs are required to follow the supplied recording, including its speech rate and pauses, while adjusting the visible speaker's articulation.

\textbf{Background editing.} A source video and background image specify the replacement environment. Structure-guided cases follow the reference's major structures and spatial layout; natural-adaptation cases allow their arrangement to fit the source viewpoint and lighting. Both variants preserve the foreground actors, clothing, performance, handled objects, soundtrack, and camera behavior, including actor--object contacts.

\subsection{Benchmark Construction and Statistics}
\label{sec:construction}
\begin{figure}[!t]
\centering
\includegraphics[width=0.99\linewidth]{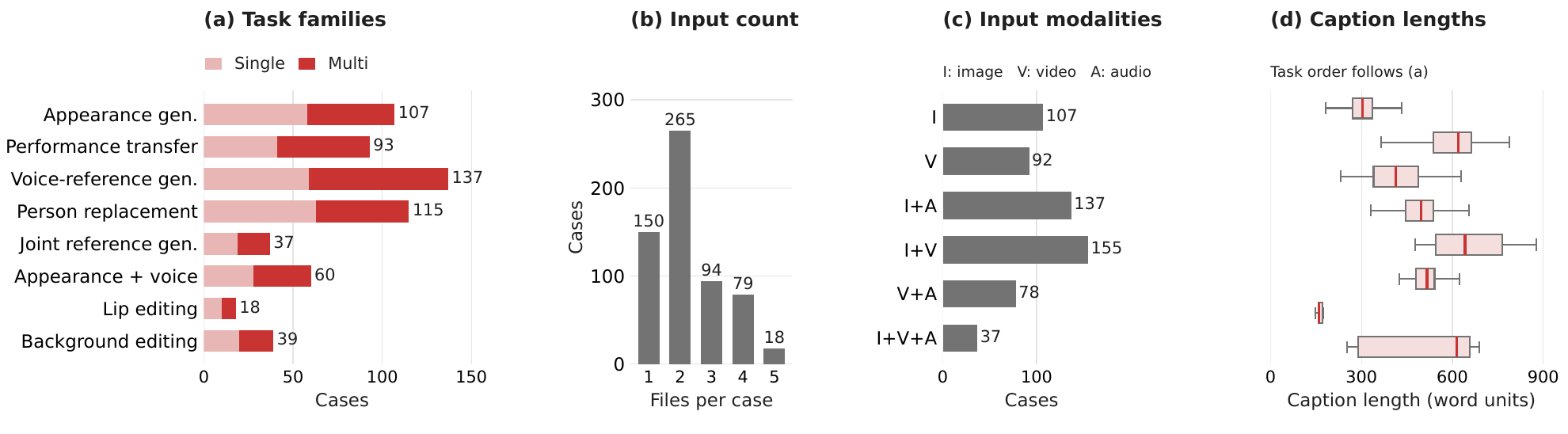}
\caption{Composition of HumanCtrl-Bench. (a) Task counts by subject setting. (b--c) Reference-file counts and modality combinations; audio embedded in a video is not an additional file. (d) Caption lengths pooled across the five models within each task; task order follows (a). Boxes show medians and interquartile ranges; whiskers reach observations within 1.5 interquartile ranges, with outlier markers omitted. Each alphanumeric word or Chinese character counts as one unit.}
\label{fig:benchmark-statistics}
\end{figure}

\textbf{Selecting materials for each role.} We assemble cases from character portraits, film and television clips, subject-linked speech samples, and source videos for editing. We use existing image crops and clips. The performance pool includes synthesized demonstrations, and the joint-editing material includes appearance-edited clips and derived dubbing. Within each candidate pool, we group entries by target video path, remove repeated targets, and sample across source titles in turn. For image-guided edits, the additional portrait or background image specifies the replacement. For externally sourced portraits and background photographs, the collection records attribution and available license metadata. A clip can support several instructions, so the unit counted in the benchmark is a task instance.

\textbf{Binding references and dialogue to actors.} For each case, the reference list records each asset's role and position, the associated subject where applicable, and whether a video supplies audio. We link an appearance image to its target actor and a performance-video actor to the person who should reproduce the action. We select a speech segment for each designated timbre reference, checking ownership against the donor interval when available. The caption assigns each spoken line to a speaker; silent participants and off-screen voices retain their own roles. This distinction matters even when both visible actors have voice samples: either actor may remain silent. In paired dialogue cases, the two instructions use identical ordered reference files and actor assignments but request different words. For the 18 lip-editing cases the audio input instead supplies the complete recording, including its timing, alongside a source video.

\begin{figure}[!t]
\centering
\includegraphics[width=\linewidth]{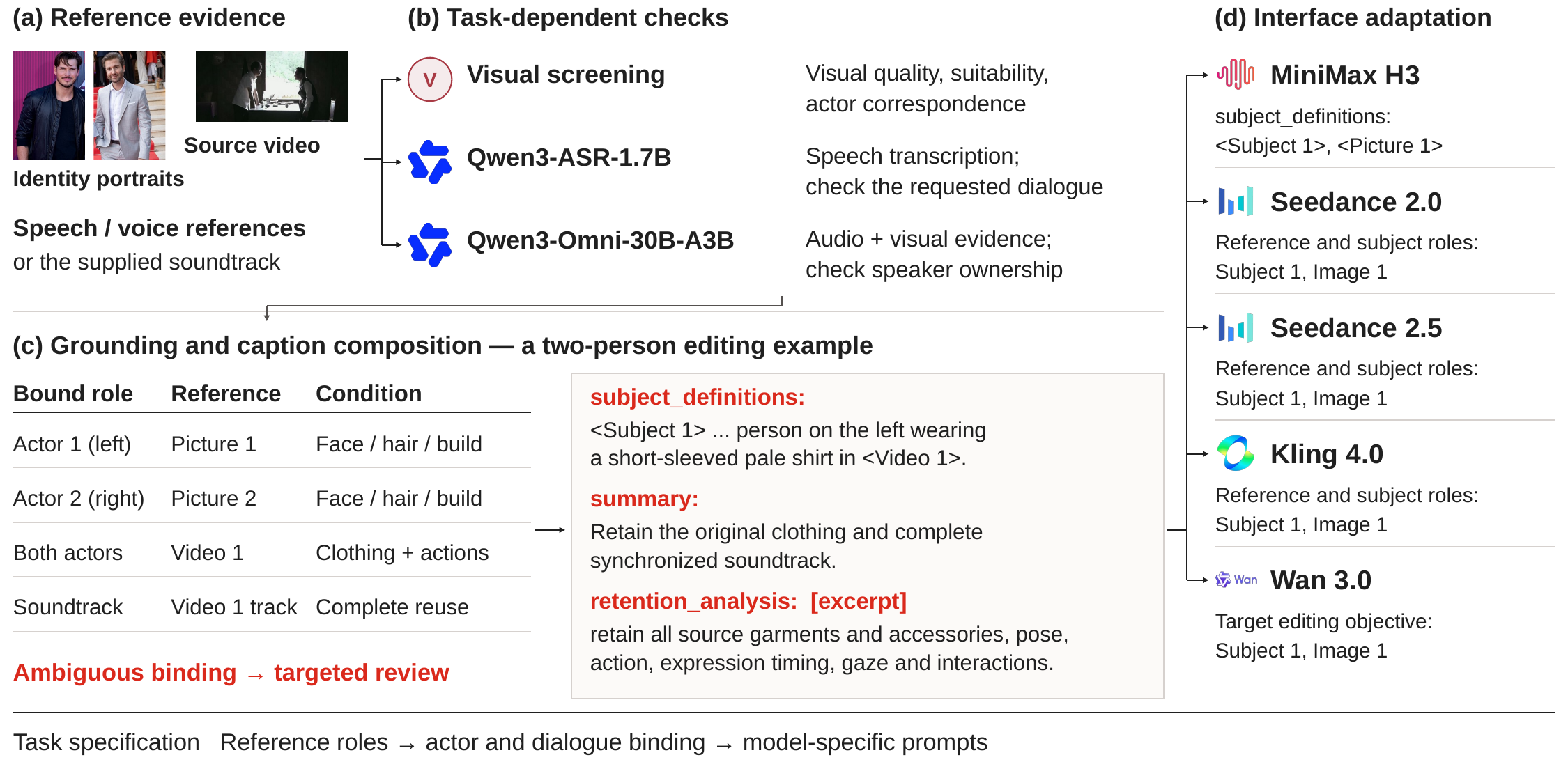}
\caption{Caption construction and adaptation to five generation interfaces. (a--b) Reference evidence undergoes task-dependent visual, transcription, and speaker-ownership checks. (c) A two-person edit binds each identity portrait to a source actor and separates transferred attributes from retained content; the text shows MiniMax H3~\citep{minimax2026h3} caption excerpts. (d) Interface-specific labels and instruction structures express these requirements.}
\label{fig:caption-construction}
\end{figure}

\textbf{Adapting captions to model interfaces.} Figure~\ref{fig:caption-construction} summarizes the information flow from role-labeled references to model-specific instructions. Each instruction describes the scene or edit, the role of each reference, the target dialogue, and the attributes to preserve. Model-specific captions express these requirements using the interface's media labels, reference order, subject declarations, and dialogue syntax. The English descriptions retain quoted speech, lyrics, and visible text in their original language. We store the submitted request and its input mapping alongside the exported caption and generated clip. Evaluation uses the reference roles, actor assignments, and editing scopes in the benchmark task specifications; saved requests and input mappings retain the generation provenance.

\textbf{Screening and verification.} Candidate review checks visual clarity, suitability for the specified task, actor correspondence, and the availability of required reference materials. Voice review separately checks speaker ownership, intelligibility, and interference. Targeted human review addresses flagged references, ambiguous actions, and dialogue assignments. Final checks verify correspondence between references and task descriptions and the availability of outputs from all five models for each case. For the 39 background edits, the audit also checks all 195 model captions against the saved generation prompts and distinguishes structure guidance from natural adaptation.

\textbf{Benchmark coverage.} HumanCtrl-Bench contains 606 cases: 374 generation and 232 editing, with 298 single-subject and 308 multi-subject cases. Figure~\ref{fig:benchmark-statistics}(a--c) and Table~\ref{tab:tasks} give the family counts, reference-file combinations, and editing scopes. The 3,030 generated outputs give one observation per system and case.

\textbf{Prompt statistics.} Figure~\ref{fig:benchmark-statistics}(d) pools caption lengths across the five models. Appendix Figure~\ref{fig:prompt-distributions} separates caption lengths by task and model, counting alphanumeric words and individual Chinese characters. Joint-reference captions have higher median lengths than appearance-generation and lip-editing captions across all five models. MiniMax H3~\citep{minimax2026h3} also has a lower median length than the other models for lip editing.

\subsection{Evaluation Protocol}
\label{sec:evaluation}
We evaluate requested control and output quality through three complementary views. Condition measurements compare outputs with the specified targets and protected source content; MLLM judgments provide task-conditioned checklist scores; and quality estimators characterize appearance, motion, and audio production. The first view follows the task's applicable conditions, while the quality views describe the generated media. Each view retains its own scale and evaluation cohort.

\subsubsection{Task-condition measurements}
\label{sec:condition-method}
\textbf{Applicability and evidence states.} For case $i$, system $m$, metric $k$, and measurement item $r$, let $a_{imkr}=1$ indicate a required measurement under the task variant, caption, reference roles, and verified generation inputs. The required item set is $\mathcal R_{imk}=\{r:a_{imkr}=1\}$, with items defined at the video, actor, turn, or endpoint level. Appearance-reference generation uses visual reference measurements; voice-reference generation uses speech and timbre measurements. Person replacement distinguishes text-defined attributes from image-defined identity and protected source clothing. Combination tasks inherit their specified conditions. Evidence states distinguish measured, partially measured, unavailable, inapplicable, content-gated, and explicitly deferred items; video-level completeness is checked against the required item set.

\textbf{Actor correspondence and measurement targets.} The evaluation records reference people, output tracks, action endpoints, and speech turns separately. Identity matching uses one-to-one track assignment, with distinct assignment and scoring frames in the cross-fitting branch; independent person/garment matching estimates appearance-set similarity. Pose measurements use the task-designated performer and target role to select endpoint keypoints. Speech measurements bind reference audio to dialogue roles and concatenate selected same-role segments. Ambiguous assignments remain uncertain, and silence or off-screen speech follows the task mask. Appendix~\ref{sec:binding-audit} gives concrete face-ownership and endpoint-correspondence checks. Appearance resemblance, action assignment, and dialogue ownership retain separate interpretations.

\textbf{Visual conditions.} Target identity uses ArcFace~\citep{deng2019arcface}; segmented-person and garment appearance use DINO~\citep{caron2021dino}. Text-directed edits use whole-person, face, hair, and garment CLIP~\citep{radford2021clip}. Background editing reports target and source background similarities, their difference, source-person identity, and illumination compatibility (Harmony) computed with BargainNet~\citep{BargainNet2021} separately. Positive target-minus-source similarity expresses relative resemblance; natural-adaptation instructions permit changes in layout and viewpoint. Performance transfer uses caption-selected endpoint pose features after duplicate-box consolidation and target-role binding. An unavailable endpoint can use the nearest valid frame within 0.5\,s inward, retaining the temporal offset. The mapped 0--100 pose value measures endpoint geometry, while target-identity evidence checks whether the requested person was transferred.

\textbf{Speech conditions.} Qwen3-ASR~\citep{shi2026qwen3asrtechnicalreport} supplies transcripts for computing character error rate (CER) and word error rate (WER); Qwen3-ForcedAligner~\citep{shi2026qwen3asrtechnicalreport} supplies timing anchors. ECAPA-TDNN~\citep{desplanques2020ecapa,speechbrain} compares role-bound generated speech with its assigned reference using raw cosine similarity. Reliably separable same-role segments are concatenated. Short speech, single-unit responses, and local phrases are retained with their scope and uncertainty flags. Difficult cases can use manually confirmed timing or actor assignments, recorded with the measurement. WER strictly above 50\% gates subsequent timbre evaluation and both lip-editing measurements. Previously measured gated timbre remains diagnostic. Explicitly required but confirmed absent speech is a condition failure, recorded separately from a measured cosine.

For supplied-recording lip editing, the opening offset is the signed mean generated-minus-reference start time of the first three aligned characters or words. The records retain the actual anchor count and position for shorter or skipped-opening fallbacks (Appendix~\ref{sec:condition-details}). Speaker-bound SyncNet~\citep{Chung16a} evaluates the selected longer utterances within the generated video; terminal short replies are outside the evaluated scope. Opening displacement relative to the reference recording and audiovisual lag within the generated video are distinct quantities.

\textbf{Common-case aggregation.} Let $b_{imk}=1$ when $\mathcal R_{imk}$ is nonempty, the metric record is marked measured or accepted nearby-frame recovery, every required item has a numeric value, and no content gate or confirmed speech-absence failure applicable to metric $k$ is triggered. The metric value $x_{imk}$ follows its recorded aggregation: mean role cosine for timbre, the minimum required endpoint score for pose, and signed opening displacement for timing. Measurements with shortened temporal support retain scope flags. When word segmentation leaves WER unavailable, timbre values retain an unknown-gate flag. For task or task--subject group $g$, the common-case set and mean are
\begin{equation}
\mathcal C_{gk}=\{i\in g:\ \forall m,\ b_{imk}=1\},\qquad
\mu_{gmk}=\frac{1}{|\mathcal C_{gk}|}\sum_{i\in\mathcal C_{gk}}x_{imk}.
\label{eq:condition-common}
\end{equation}
The mean is reported when $\mathcal C_{gk}$ is nonempty. Each case receives equal weight within a metric. Main-table rows pool subject groups within a task; Appendix~\ref{sec:condition-details} separates all 16 task--subject settings. CER/WER are macro means of case-level error rates and can exceed 100\% through insertions. Each row reports its native scale and common/applicable case counts: raw cosines and CLIP/Harmony values, macro CER/WER percentages, a 0--100 endpoint-pose proxy, and millisecond timing offsets. Opening offsets are signed (positive means later speech); synchronization uses absolute lag. Appendix~\ref{sec:condition-details} quantifies narrowed measurement support. Missing values remain unfilled, and the resulting comparisons form a profile of individual conditions.

\textbf{Overall mean rank.} We aggregate task-condition evidence through within-case pairwise comparisons. Metric differences are transformed using fixed directions, scales, and tolerances, then combined using task weights; unavailable values retain contribution bounds. Cases with all ten resolved model pairs receive ranks for all five models from win/tie/loss credits of 1, 0.5, and 0. We average these ranks within each of the eight tasks' single-/multi-person groups, form tied group positions using a 0.20 tolerance, and average positions equally across groups with rankable cases. The last row of Table~\ref{tab:condition-main} reports this descriptive mean rank; Appendix~\ref{sec:condition-ranking} gives weights, formulas, and group coverage.

\subsubsection{Model-based checklist judgments}
MiMo-V2.6-Flash-MOPD~\citep{xiaomi2026mimoflashmopd} evaluates one generated candidate per request using its task caption, role-bound references, and applicable checklist. It directly receives ordered visual panels sampled at 8\,fps and the applicable raw audio. Caption-blind transcriptions from MiMo supply the dialogue gate; no external auditory expert is used. Each request uses a fresh context without other candidates or prior pairwise judgments. The evaluator returns a task-completion score and separate applicable condition scores on an ordinal 0--4 scale, from an absent or contradicted requirement to all observable requirements satisfied. The task score is assessed from the specified conditions rather than obtained by adding condition scores or generic quality penalties.

Table~\ref{tab:mimo-checklist} reports sample-equal means on the cases scored for all five models within each task. Uncertain, inapplicable, gated, and deferred entries retain separate states and are not filled with zero. For the overall rank, we rank the five task scores within each common case, average sample ranks within task--subject groups, form tied group positions using the same 0.20 mean-rank tolerance as Table~\ref{tab:condition-main}, and average the positions equally across the 14 eligible groups. Appendix~\ref{sec:checklist-details} gives the score anchors, gating rules, and rank-aggregation formulas.

Precise opening alignment and lip synchronization remain deferred unless gated by dialogue failure; the lip-editing family is excluded from the overall summaries.

\subsubsection{Audiovisual quality}
\textbf{ Visual quality.} We assess image/video quality and aesthetics with MUSIQ-SPAQ~\citep{ke2021musiq}, DOVER++~\citep{wu2023dover}, Aesthetic Predictor V2.5~\citep{aestheticpredictor25}, and the LAION aesthetic head~\citep{laionaesthetic}. Temporal measurements use AMT-S~\citep{li2023amt} interpolation-based smoothness, adjacent-frame consistency using CLIP~\citep{radford2021clip} and DINO~\citep{caron2021dino}, and VBench~\citep{huang2024vbench} subject/background consistency; dynamic degree describes motion quantity.

\textbf{ Audio quality.} Audiobox Aesthetics~\citep{tjandra2025aes} measures audio production quality. Appendix~\ref{sec:quality-details} reports the quality estimators' scales and results.

\section{Experiments and Diagnostic Analysis}
\label{sec:experiments}

\subsection{Experimental scope}
To assess multi-condition control across commercial and open-weight audiovisual models, we evaluate Seedance~2.0~\citep{seedance2026seedance20advancingvideo}, Seedance~2.5~\citep{bytedanceseed2026seedance25}, Kling 4.0~\citep{kuaishou2026kling40}, and Wan 3.0~\citep{alibabacloud2026wan30} alongside MiniMax H3~\citep{minimax2026h3}, which accepts image, video, and audio references. All five models are evaluated on the same 606 cases, with one output per system and case. Prompt formatting follows the model-specific adaptation procedure in Section~\ref{sec:construction}. We first examine which control tasks favor each system, then assess task-conditioned MLLM judgments and output quality. Selected edits finally identify the source attributes that can change unintentionally.

\subsection{Condition-specific control and coverage}
\textbf{Identity, action, and preservation have different profiles.} Table~\ref{tab:condition-main} compares systems within each requested condition. Wan 3.0 has the highest target-identity cosine on 90 common performance-transfer cases (0.825), while MiniMax H3 has the highest endpoint-pose value on the 49 commonly measurable pose cases (54.3). These metric-specific cohorts assess complementary parts of performance transfer. Image-conditioned person replacement likewise separates target identity from source-clothing retention: Wan 3.0 leads identity cosine (0.696 on 23 cases), whereas Seedance~2.5 leads source-garment similarity (0.916 on 22). For background editing, Kling 4.0 has the largest target-minus-source background difference (0.357 on 35), while Seedance~2.5 has the highest retained source-identity similarity (0.969 on 31).

\input{tables/condition_main.tex}

\textbf{Group-balanced task-condition rank.} On 503 cases with ranks for all five models across all 16 task/subject groups, Wan 3.0 ranks first overall, with a mean rank of 1.625, followed by Seedance~2.0 (3.000) and Seedance~2.5 (3.250). Each group receives equal weight under the fixed comparison weights and tolerances. A case is rankable when every pair outcome is determined, including by dialogue gates or missing-evidence bounds; the component rows instead require complete in-scope numeric records for each metric. The rank and component means therefore use different case sets. Appendix Figure~\ref{fig:condition-task-profiles} shows the task-wise single-/multi-subject rank profiles and their coverage.

\textbf{Timbre and dialogue test separate voice conditions.} Wan 3.0 has the highest raw timbre cosine on the 131 common voice-reference cases (0.622), while Kling 4.0 has the lowest macro CER on all 137 cases (2.02\%). The subject-count breakdown preserves this distinction: Wan 3.0 leads timbre in both groups, whereas the lowest CER belongs to Seedance~2.5 for single-subject cases and Kling 4.0 for multi-subject cases (Appendix Table~\ref{tab:condition-subject}). Group profiles describe their respective case sets. Across the three voice-conditioned families, all required timbre entries have a recorded measurement or failure disposition; the appendix separates numeric cosines, speech-absence failures, dialogue gates, and partial speech support.

\textbf{Coverage determines the reach of each comparison.} Complete endpoint-pose proxies are available for 303/410 applicable videos (73.9\%); 48 have partial endpoint evidence and 59 have none. For lip editing, 29/90 videos fail the dialogue gate. All 61 remaining videos have opening offsets, and 73 of 80 in-scope long turns have synchronization estimates. Seven long turns remain unavailable; 16 terminal short replies are outside the selected synchronization scope. No lip-editing case has ungated timing evidence for all five models. Timing results are therefore reported on each system's available cases in Appendix~\ref{sec:condition-details}, together with anchor support and the full coverage summary.

\begin{figure}[!t]
\centering
\includegraphics[width=0.93\linewidth]{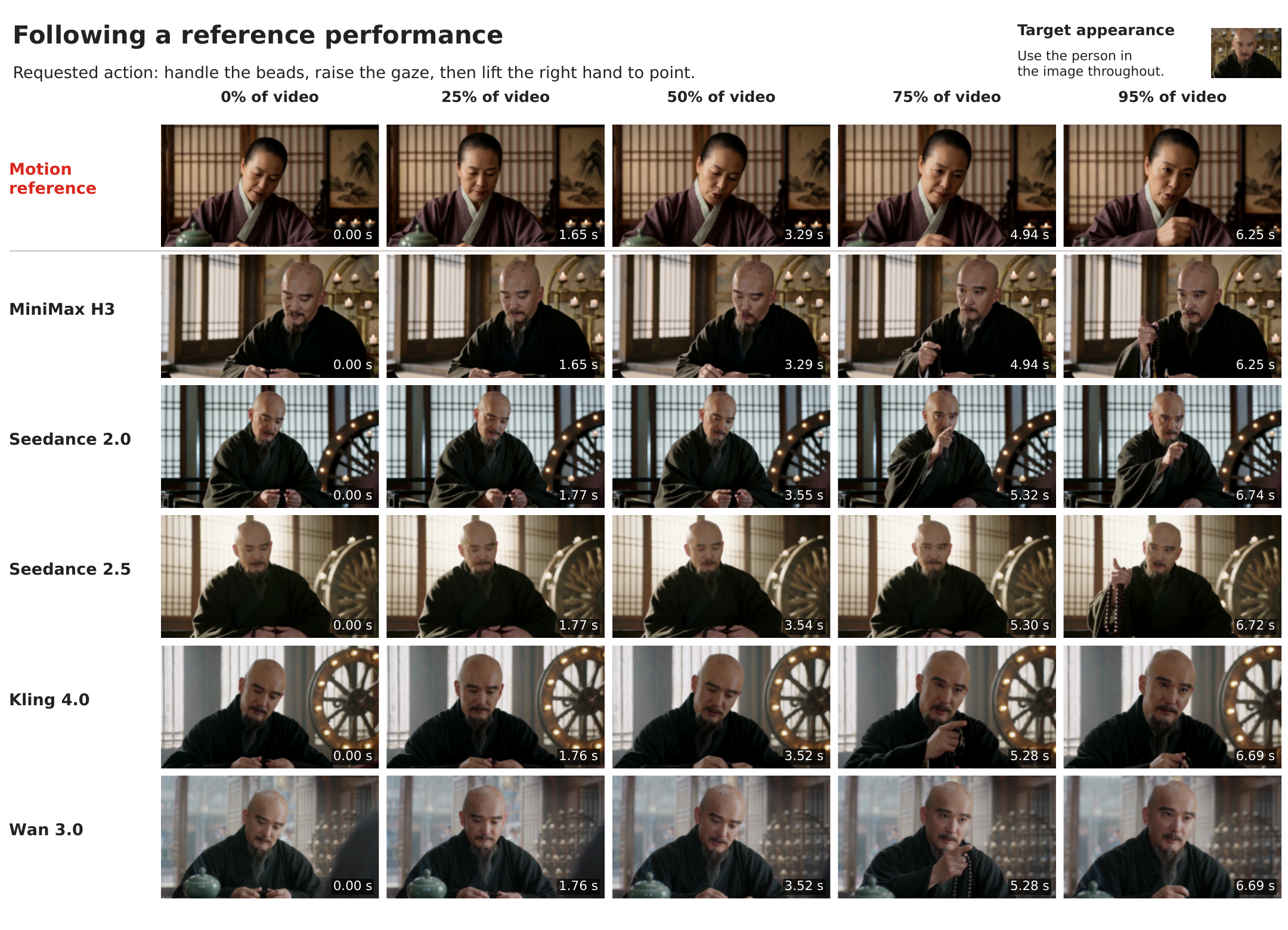}
\caption{Reference-performance following across five models on one appearance-conditioned motion-transfer task. The target person is specified by the portrait; the reference performance progresses from handling beads with a lowered gaze to raising the gaze and right hand. Columns sample 0\%, 25\%, 50\%, 75\%, and 95\% of each saved video's duration; insets show actual timestamps. Panels use the same 16:9 center crop without stretching. The columns compare relative progress, without motion-based temporal alignment.}
\label{fig:motion-following}
\end{figure}

\textbf{Following the reference performance over time.} Figure~\ref{fig:motion-following} compares five models on a task that transfers bead handling and a pointing gesture to a specified person. At 75\% of each video's duration, Seedance~2.0, Kling 4.0, and Wan 3.0 show a raised right hand, while Seedance~2.5 still holds its hands low. At 95\%, Seedance~2.5 shows the raised gesture, whereas Kling 4.0 and Wan 3.0 have lowered their hands. These sampled states expose differences in gesture timing and duration that an endpoint-pose summary alone does not describe.

\subsection{Model-based checklist task assessment}
Table~\ref{tab:mimo-checklist} reports MiMo judgments on 513 common cases across 14 task--subject groups. Seedance~2.5 obtains the second-highest voice-reference score (3.903), while Kling 4.0 leads this task (3.938). Wan 3.0 leads appearance generation, person replacement, joint appearance--voice editing, and background editing, and ranks first overall by mean rank (2.679). MiniMax H3 has the highest performance-transfer score (3.253). MiMo directly assesses audiovisual inputs.

\begin{table}[!t]
\centering
\caption{Independent audiovisual checklist scores from MiMo-V2.6-Flash-MOPD~\citep{xiaomi2026mimoflashmopd} (0--4). $C/N$ denotes cases scored for all five models / task cases. Overall mean rank averages group positions equally over 14 task--subject groups after within-case ranking. Seven task families contribute 513 common cases. Bold and underline mark the best and second-best displayed values, including ties.}
\label{tab:mimo-checklist}
\fontsize{8.5}{10}\selectfont
\setlength{\tabcolsep}{3pt}
\renewcommand{\arraystretch}{1.10}
\begin{tabular*}{\linewidth}{@{\extracolsep{\fill}}lrrrrrr@{}}
\arrayrulecolor{jingdongred}\toprule\arrayrulecolor{black}
\paperth{Task} & \paperth{$C/N$} & \paperth{MiniMax H3} & \paperth{Seedance 2.0} & \paperth{Seedance 2.5} & \paperth{Kling 4.0} & \paperth{Wan 3.0} \\
\midrule
Appearance & 106/107 & 3.519 & \underline{3.670} & 3.585 & 3.604 & \textbf{3.708} \\
Performance & 91/93 & \textbf{3.253} & \underline{3.176} & 3.165 & 3.165 & 3.121 \\
\begin{tabular}[t]{@{}l@{}}Voice\\reference\end{tabular} & 113/137 & 3.735 & 3.832 & \underline{3.903} & \textbf{3.938} & 3.655 \\
\begin{tabular}[t]{@{}l@{}}Person\\replacement\end{tabular} & 102/115 & 3.157 & 3.265 & 3.118 & \underline{3.343} & \textbf{3.441} \\
\begin{tabular}[t]{@{}l@{}}Joint\\reference\end{tabular} & 24/37 & 3.000 & \textbf{3.375} & \underline{3.208} & \textbf{3.375} & 2.792 \\
\begin{tabular}[t]{@{}l@{}}Appearance\\+ voice\end{tabular} & 49/60 & 3.000 & \underline{3.061} & 2.918 & 3.000 & \textbf{3.143} \\
Background & 28/39 & 2.071 & 2.893 & \underline{3.464} & \underline{3.464} & \textbf{3.500} \\
\midrule
\paperth{\begin{tabular}[t]{@{}l@{}}Overall mean\\rank $\downarrow$\end{tabular}} & 513/606 & 3.536 & 3.036 & 3.000 & \underline{2.750} & \textbf{2.679} \\
\bottomrule
\end{tabular*}
\end{table}

\subsection{Objective audiovisual quality}
Table~\ref{tab:quality-diagnostics} evaluates all 606 cases using quality estimators on the scales indicated in the table. Kling 4.0 has the highest displayed MUSIQ, Aesthetic Predictor V2.5, and Audiobox production-quality means. Seedance~2.0 has the highest adjacent-frame CLIP/DINO and subject-consistency means. The estimators therefore describe different output properties. For example, Seedance~2.0 combines the highest adjacent-frame consistency with the lowest dynamic degree (25.083\%). Production quality measures a separate property from reference-voice or recording retention.

\begin{table}[!t]
\centering
\caption{Audiovisual quality estimates on 606 common cases per system, with equal case weights. Metrics use the scales and transformations shown in the row labels. Higher is better; dynamic degree describes motion quantity and is not ranked. Bold and underline mark the best and second-best displayed means, including ties.}
\label{tab:quality-diagnostics}
\fontsize{8.5}{10}\selectfont
\setlength{\tabcolsep}{4pt}
\renewcommand{\arraystretch}{1.10}
\begin{tabular*}{\linewidth}{@{\extracolsep{\fill}}lrrrrr@{}}
\arrayrulecolor{jingdongred}\toprule\arrayrulecolor{black}
 \paperth{Measurement} & \paperth{MiniMax H3} & \paperth{Seedance 2.0} & \paperth{Seedance 2.5} & \paperth{Kling 4.0} & \paperth{Wan 3.0} \\
\midrule
\rowcolor{tablesection}
\multicolumn{6}{@{}l}{\textbf{Visual quality and aesthetics}} \\
MUSIQ-SPAQ & 57.555 & \underline{63.152} & 56.985 & \textbf{65.762} & 60.728 \\
DOVER++ technical $\times100$ & -0.429 & 0.237 & -0.935 & \textbf{1.623} & \underline{0.588} \\
DOVER++ aesthetic (raw) & 0.054 & 0.059 & 0.051 & \textbf{0.066} & \underline{0.064} \\
Aesthetic Predictor V2.5 & 4.591 & 4.683 & 4.703 & \textbf{5.071} & \underline{4.865} \\
LAION aesthetic /10 & 0.573 & \textbf{0.600} & 0.576 & \underline{0.594} & 0.583 \\
\midrule
\rowcolor{tablesection}
\multicolumn{6}{@{}l}{\textbf{Temporal consistency and motion}} \\
AMT-S smoothness & \textbf{0.996} & \textbf{0.996} & \textbf{0.996} & \textbf{0.996} & \underline{0.995} \\
DINO adjacent-frame consistency & 0.960 & \textbf{0.970} & \underline{0.967} & 0.964 & 0.960 \\
CLIP adjacent-frame consistency & 0.970 & \textbf{0.975} & \underline{0.972} & 0.971 & 0.968 \\
Subject consistency & 0.960 & \textbf{0.967} & 0.965 & \underline{0.966} & 0.960 \\
Background consistency & 0.964 & 0.961 & \underline{0.968} & \textbf{0.969} & 0.964 \\
Dynamic degree (\%, descriptive) & 37.954 & 25.083 & 26.403 & 32.838 & 34.818 \\
\midrule
\rowcolor{tablesection}
\multicolumn{6}{@{}l}{\textbf{Audio quality}} \\
Audiobox production quality & 7.046 & \underline{7.074} & 6.773 & \textbf{7.287} & 7.068 \\
\bottomrule
\end{tabular*}
\end{table}

\textbf{Reading the three evaluations together.} Kling 4.0 leads several visual and audio production quality estimators (Table~\ref{tab:quality-diagnostics}). In voice-reference generation, Wan 3.0 leads timbre similarity and Kling 4.0 has the lowest CER. In the MiMo checklist evaluation, Wan 3.0 ranks first by overall mean rank (Table~\ref{tab:mimo-checklist}). Each comparison retains its reported cohort and scale. Appendices~\ref{sec:quality-details} and~\ref{sec:attribute-details} detail quality and preservation measurements.

\begin{figure}[!t]
\centering
\includegraphics[width=0.84\linewidth]{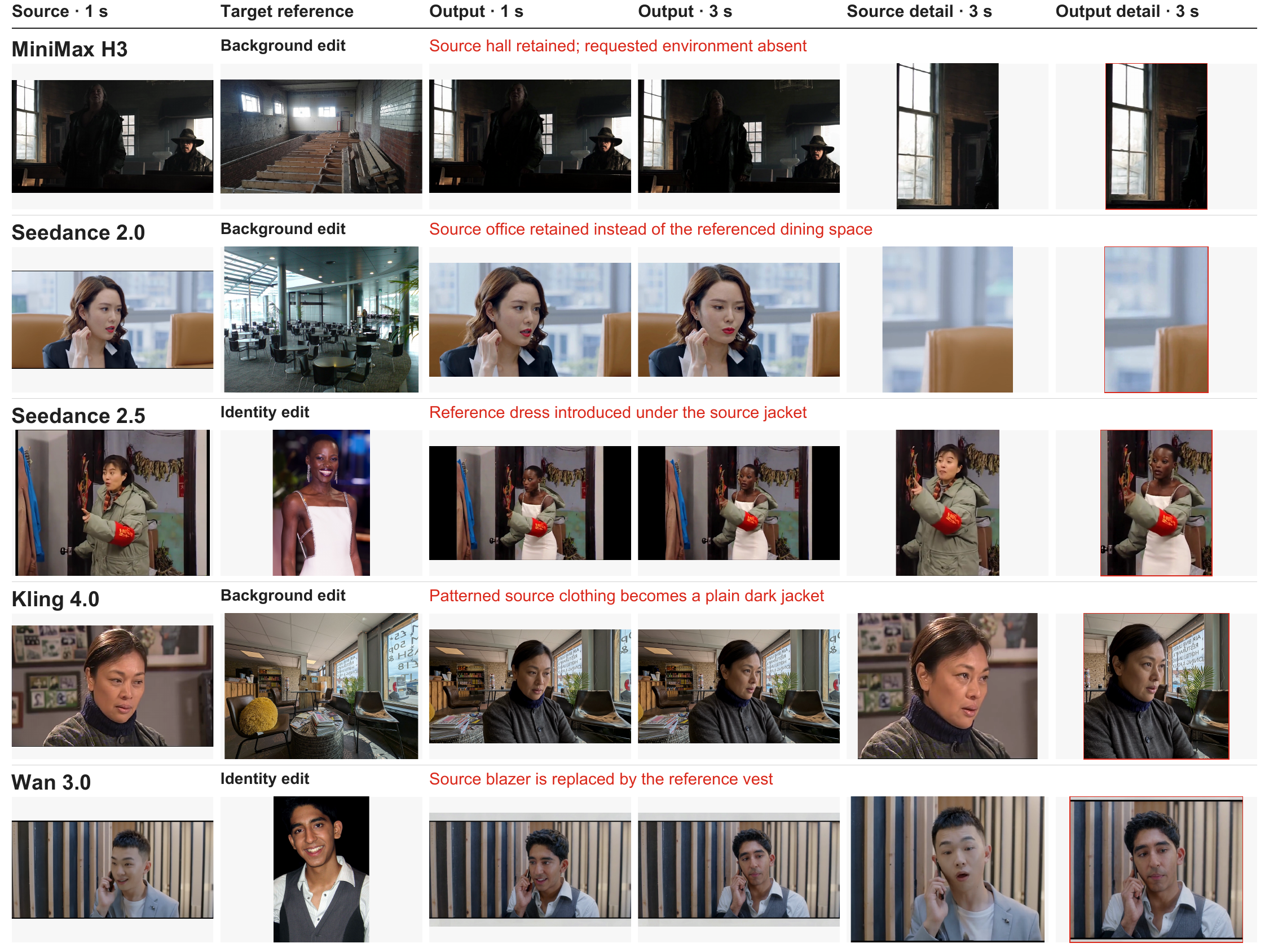}
\caption{Omitted edits and unintended clothing changes in sampled frames from five distinct cases. MiniMax H3~\citep{minimax2026h3} and Seedance 2.0~\citep{seedance2026seedance20advancingvideo} retain the source background; Seedance 2.5~\citep{bytedanceseed2026seedance25} and Wan 3.0~\citep{alibabacloud2026wan30} introduce reference clothing during identity editing; Kling 4.0~\citep{kuaishou2026kling40} changes foreground clothing during background editing. Outputs at 1 and 3\,s are paired with source/reference inputs, and the last two columns compare source/output details at 3\,s. All panels preserve aspect ratio; person details retain the face and clothing.}
\label{fig:preservation-failures}
\end{figure}

\newpage
\subsection{Model failures in selective human control}
Figure~\ref{fig:preservation-failures} shows five distinct cases spanning background and identity editing. In the displayed frames, MiniMax H3 and Seedance~2.0 retain the original environments instead of introducing the requested replacements. Seedance~2.5 introduces a reference dress beneath the source jacket, and Wan 3.0 replaces a source blazer with the reference vest. In a background edit, Kling 4.0 changes patterned foreground clothing into a plain dark jacket. The task instructions require the intended replacement while preserving source clothing. These examples illustrate both omitted changes and changes outside the requested scope.

\section{Conclusion}
HumanCtrl-Bench organizes audiovisual generation and editing around attribute transfer, actor binding, and source preservation. In voice-reference generation, Wan 3.0 had the highest timbre similarity on 131 common cases, while Kling 4.0 had the lowest character error rate on all 137 cases (Table~\ref{tab:condition-main}). The inspected edits make the preservation requirement concrete, exposing changes to clothing and accessories that should remain fixed. Role-labeled conditions, traceable requests, and metric-specific coverage provide a basis for evaluating both what a model changes and what it keeps.

\bibliographystyle{abbrvnat}
\bibliography{references}

\clearpage
\appendix
\section{Task Definitions and Controlled Contrasts}
\label{sec:variants}
Table~\ref{tab:variants} details the control tasks within each family. Single and multi refer to the controlled or referenced principal actors. The full task description also records input roles, actor assignments, and the content to retain.

\begin{table}[!t]
\caption{Eleven fine-grained tasks and their single-/multi-subject settings. S and M count cases by principal-actor setting; N gives the total.}
\label{tab:variants}
\centering

\footnotesize
\setlength{\tabcolsep}{3pt}
\begin{tabular}{p{0.20\linewidth}p{0.51\linewidth}rrr}
\arrayrulecolor{jingdongred}\toprule\arrayrulecolor{black}
 \paperth{\centering\arraybackslash Family} & \paperth{\centering\arraybackslash Control task} & \cth{S} & \cth{M} & \cth{N} \\
\midrule
Appearance generation & Image-referenced appearance generation & 58 & 49 & 107 \\
\paperrowsep
Performance transfer & Identity-conditioned performance transfer & 41 & 52 & 93 \\
\paperrowsep
Voice-reference generation & Reference-voice generation with specified dialogue & 59 & 78 & 137 \\
\paperrowsep
Appearance editing & Face replacement with source-clothing retention & 9 & 10 & 19 \\
\paperrowsep
Appearance editing & Identity, hairstyle and visible-build replacement & 2 & 2 & 4 \\
\paperrowsep
Appearance editing & Text-directed appearance editing with source audio & 52 & 40 & 92 \\
\paperrowsep
Joint reference generation & Joint appearance, performance and voice control & 19 & 18 & 37 \\
\paperrowsep
Appearance and voice editing & Text-directed appearance and reference-voice editing & 28 & 32 & 60 \\
\paperrowsep
Supplied-recording lip editing & Lip editing with a retimed supplied recording & 10 & 8 & 18 \\
\paperrowsep
Background edit & Background replacement with natural adaptation & 10 & 10 & 20 \\
\paperrowsep
Background edit & Background replacement with structural guidance & 10 & 9 & 19 \\
\midrule
Total & Eleven tasks, 22 task/subject settings & 298 & 308 & 606 \\
\bottomrule
\end{tabular}
\end{table}

\subsection{Source-audio conditions}
The reference manifest specifies that video audio is not used as conditioning in the 130 performance-transfer and joint-reference generation cases. It specifies source-audio use in 154 editing cases that retain the original soundtrack: face replacement, broader identity replacement, text-directed appearance editing, and background editing. The sixty appearance and voice edits pair a source video with separate timbre samples; the eighteen supplied-recording lip edits pair a source video with a complete recording. These assignments describe audio-conditioning roles rather than whether the source files contain an audio track.

The benchmark task specification defines the reference roles, dialogue ownership, and source-preservation requirements used for evaluation. Appearance editing requires retention of the complete source soundtrack, including dialogue, ambience, effects, and music. A designated off-screen acknowledgment remains an independent voice: the visible listening actor stays silent and its available timbre sample is unused. Source audio, timbre references, and off-screen speech are therefore assigned separately in the applicable measurement mask.

\subsection{Selective appearance and background edits}
The nineteen face-replacement cases retain source clothing, performance, setting, camera, and soundtrack. The four broader identity-editing cases additionally transfer visible hairstyle and build, while retaining source clothing, headwear, accessories, and the complete soundtrack. Hair and build transfer follow the regions visible in both the source and reference. The ninety-two text-directed appearance cases retain the source soundtrack.

All thirty-nine background edits supply source audio and a background image. The nineteen reference-structure cases use the depicted structures, furnishings, materials, and spatial layout to specify the new environment. The twenty natural-adaptation cases permit changes to layout and object details for integration with the source camera and lighting. Both variants retain the foreground people, their clothing and performance, handled objects, and source soundtrack.

\subsection{Dialogue, actor assignment, and recording timing}
Ten voice-reference pairs keep their ordered references and speaker-to-timbre assignments fixed while comparing an existing task script with newly specified dialogue. These designated pairs assess dialogue changes under the same reference conditions.

The eighteen supplied-recording lip-editing cases form nine early/late pairs with a 0.6-second offset difference. Each pair retains the same recorded content and speech rate. Full-recording reuse requires the output to preserve the supplied speech rate and pauses; it is not permission to regenerate the words at a different pace. Voice-reference generation instead conditions on timbre and follows the target dialogue and any explicit delivery instructions, without a general requirement to copy the reference speech rate. The task supplies a source video and a complete recording, retains non-mouth identity, clothing, gaze, body motion, and camera, and requests lip and jaw articulation for the supplied speech and pauses. The reported opening-offset and selected-turn synchronization measurements assess timing on their stated support; they do not establish preservation of the complete recording or its full speech-rate trajectory.

Table~\ref{tab:diagnostic-strata} summarizes overlapping analysis subsets.

\begin{table}[!t]
\caption{Diagnostic subsets for reference count, source audio, dialogue, recording timing, and editing scope. The subsets overlap within the 606-case suite.}
\label{tab:diagnostic-strata}
\centering

\small
\setlength{\tabcolsep}{4pt}
\begin{tabular}{p{0.24\linewidth}p{0.66\linewidth}}
\arrayrulecolor{jingdongred}\toprule\arrayrulecolor{black}
 \paperth{\centering\arraybackslash Diagnostic facet} & \paperth{\centering\arraybackslash Coverage} \\
\midrule
Subject setting & 298 single-subject and 308 multi-subject cases; all eleven tasks contain both groups \\
\paperrowsep
Reference cardinality & One reference: 150 cases; two: 265; three: 94; four: 79; five: 18 \\
\paperrowsep
Source-audio conditioning & Enabled in 154 source-based cases; separate audio references in 78 \\
\paperrowsep
Dialogue relation & Ten new-dialogue and ten existing-script cases sharing references in ten pairs \\
\paperrowsep
Soundtrack timing & Nine early and nine late cases; nine paired offsets separated by 0.6 seconds \\
\paperrowsep
Appearance-edit scope & Text-directed appearance 92; face replacement 19; identity/hair/build replacement 4 \\
\paperrowsep
Background guidance & Reference-structure guidance 19; natural adaptation 20 \\
\bottomrule
\end{tabular}
\end{table}

\section{Task-Condition Measurements and Coverage}
\label{sec:condition-details}
Measurements cover the benchmark dataset: 606 cases, 298 single-subject and 308 multi-subject, and 3,030 outputs. Coverage is computed for each video from its applicable roles and dialogue turns. Table~\ref{tab:condition-coverage} distinguishes applicable evidence from measurements that are unavailable, partial, gated, or inapplicable. F denotes numeric measurements without an explicit partial-coverage flag. N includes conditions not specified as applicable. The 16 deferred synchronization turns occur in 11 videos; this scope rule removes no entire video. Tables~\ref{tab:condition-subject} and~\ref{tab:lip-available} report subgroup and available-case summaries, respectively.

\subsection{Actor correspondence and measurement audits}
\label{sec:binding-audit}
The benchmark separates the requested actor assignment from the correspondence inferred during measurement. Table~\ref{tab:binding-evidence} states the evidence available for each component. Reference-to-track assignment identifies a candidate person; action correspondence additionally needs the task-selected performer and endpoint, and vocal correspondence needs the assigned dialogue role and reference voice. Independent appearance matching can recover a reference set even when action or speaking-role correspondence remains unresolved.

\begin{table}[!t]
\centering\footnotesize
\caption{Actor correspondence used by identity, performance, and voice measurements. Each row links a recorded assignment to the condition it supports.}
\label{tab:binding-evidence}
\setlength{\tabcolsep}{4pt}
\begin{tabular}{@{}p{.18\linewidth}p{.40\linewidth}p{.35\linewidth}@{}}
\arrayrulecolor{jingdongred}\toprule\arrayrulecolor{black}
\paperth{Component} & \paperth{Recorded correspondence} & \paperth{Interpretation} \\
\midrule
Identity / appearance & Reference person to output track; one-to-one assignment, with disjoint assignment/scoring frames in the cross-fitting branch & Identity or appearance-set fidelity; independent DINO matches can differ across regions \\
\paperrowsep
Performance & Caption-designated source performer to target role; selected endpoint keypoints and valid temporal support & Geometry of the requested role at the measured endpoint \\
\paperrowsep
Voice / dialogue & Reference audio to dialogue role; selected same-role speech segments; silence and off-screen masks & Voice fidelity on retained speech; visible-speaker association remains a separate requirement \\
\bottomrule
\end{tabular}
\end{table}

\paragraph{Targeted correspondence checks.}
A generated-endpoint audit examined 40 endpoint conditions in 29 videos. Checking caption-designated actors against frames identified interference from framed photographs, screen reflections, and non-target people. In an over-the-shoulder shot, measuring the designated speaker's wrist did not require a skeleton for the foreground listener. Corrected correspondence recovered 13 endpoint scores across nine videos using the existing keypoints; confidence thresholds and existing numerical scores were unchanged. The other 27 audited endpoints remained unavailable. These counts describe the inspected cases.

A separate appearance case shows why missing measurements must be distinguished from failed identity transfer. The detector assigned one seated actor both a full-body and an upper-body box, preventing unique face ownership. Direct face extraction on the same sampled frames recovered 110/110 valid frames for MiniMax H3 and an ArcFace cosine of 0.764. The branch accepted frames with exactly one valid face, without selecting frames by target similarity. This check establishes measurement support for the visible actor; it does not establish the action or speech assignment of other characters.

These checks establish measurement support for the inspected actors and endpoints. A benchmark-wide actor-binding accuracy would additionally require independent labels for whether each requested appearance, action, and utterance belongs to the intended visible or off-screen actor. The MLLM assignment scores below remain automatic diagnostic judgments.

\subsection{Identity, appearance, and preservation}
Person masks use SAM 2.1~\citep{ravi2024sam2}, and endpoint keypoints use Ultralytics YOLO11x-pose~\citep{yolo11_ultralytics}. Frozen extractors include InsightFace's \emph{w600k\_r50} recognizer, DINO v1 ViT-B/16~\citep{caron2021dino}, and CLIP ViT-L/14@336~\citep{radford2021clip}. Track features average valid-frame embeddings before normalization and reference comparison. Measurements use fixed-role assignment, odd/even cross-fitting, or all valid frames, depending on the measurement procedure. Cross-fitting exchanges assignment and scoring frame sets, requiring consistent assignment and at least three valid frames per scoring half. Fallback measurements accept at least one valid frame and report the frame count and measurement procedure.

Independent person/garment DINO measurement uses one-to-one appearance matching without requiring a detected face. Person and garment matches can select different tracks, so these values measure appearance-set matching; action and speech binding require separate evidence. An incomplete required role set remains partial. Highest-candidate ArcFace values that permit shared tracks are excluded from the main measurements. Visible-hair and torso-geometry probes provide separate regional diagnostics.

Background DINO excludes detected foreground and compares clean patches with target and source backgrounds, using an area-weighted mean of maximum local affinities. The difference is averaged on jointly valid frames. Source identity uses one-to-one maximum-mean assignment, then the weakest person, with an ambiguity margin of 0.10. Harmony averages $\exp(-0.04212\|h_F-h_B\|_2)$ from foreground/background BargainNet~\citep{BargainNet2021} features. Natural adaptation permits scene-layout changes. Recording diagnostics use mean-centered mono 16\,kHz signals and the shared prefix; spectra use a 512-point STFT, 256-point overlap, and $\log(1+100|X|)$. Duration error and best-lag cross-correlation are separate quantities. Shared-prefix affinity measures the compared interval.

\subsection{Voice and dialogue evidence}
Qwen3-ASR-1.7B~\citep{shi2026qwen3asrtechnicalreport} transcribes the recording without the target caption; Qwen3-ForcedAligner-0.6B~\citep{shi2026qwen3asrtechnicalreport} aligns the hypothesis. CER/WER use edit counts over normalized reference units. Ordered transcript anchors, Silero VAD~\citep{silero_vad} pauses, and retained timing annotations identify same-role speech. Reliable segments are concatenated at 16\,kHz for ECAPA-TDNN; shorter-than-one-second speech and single-unit replies are allowed. Each role cosine averages generated--reference window-pair cosines, and the video value averages the required measured roles. The raw cosine is retained without a sigmoid transform.

Among 1,576 numeric role values, 346 explicitly cover only retained turns or local phrases; 247 carry a short-generated-speech flag, 262 a short-reference flag, 30 a VAD-miss flag, 120 an approximate-boundary flag, and 15 a manually assisted timing flag. These flags overlap and include retained diagnostic measurements after dialogue failure. The flags describe recorded measurement support; temporal completeness and visible-speaker binding require independent validation.

Twenty-seven timbre-applicable videos have WER $>50\%$. Their 31 existing cosine values remain diagnostic, and two additional role entries are closed by the gate without a cosine. Two ungated videos have one explicitly required but manually confirmed absent response each. The stored condition zero for such a role is excluded from cosine means. Neither ASR omission nor VAD omission alone establishes absent speech. Of 1,170 applicable videos, 1,141 have complete numeric role records outside these gates/failures; 186 of those carry explicit partial-scope evidence. WER is unavailable for 60 timbre-applicable videos (30 voice-reference, 15 joint-reference, and 15 appearance-and-voice edits) because of mixed or unsupported word segmentation. Their raw cosines retain unknown gate status, with dialogue accuracy reported separately. The 230 silent-role entries are excluded by applicability.

\subsection{Endpoint action evidence}
Caption-selected endpoint masks define two-dimensional pose features such as relative head/wrist heights and hand-to-head/shoulder distances. Duplicate-person boxes are consolidated before binding the target actor; background portraits, reflections, and unrelated people are excluded from that binding. For reference feature $z^R_{ej}$ and generated feature $z^G_{ej}$, the endpoint proxy is
\begin{equation}
M_i=\min_{(e,j)\in\mathcal E_i}100\max\left(0,1-\frac{|z^G_{ej}-z^R_{ej}|}{\max(b_j^{\min},|z^R_{1j}-z^R_{0j}|)}\right).
\end{equation}
Feature-specific floors apply when a reference endpoint is unavailable. Missing required components leave the complete proxy unavailable. Twenty-one performance-transfer and 27 joint-reference cases have no selected endpoint, leaving 82 applicable cases (410 videos). Nearest-frame recovery selects by time within 0.5\,s inward and records the offset. There are seven first-endpoint and 29 last-endpoint nearby-frame measurements. A coordinate for an occluded or out-of-frame part is not accepted solely because a detector reports high confidence. These proxies measure the selected endpoint geometry.

\subsection{Opening timing and audiovisual synchronization}
The opening statistic uses the first dialogue turn only:
\begin{equation}
\Delta_i=\frac{1}{|\mathcal A_i|}\sum_{u\in\mathcal A_i}(t^{G}_{iu}-t^{R}_{iu}),
\end{equation}
in milliseconds, with positive values indicating later generated speech. The preferred anchor set contains the first three units (Chinese characters or language-dependent aligned words). Fallbacks use two or one unit and may begin at the second unit. Of the 61 ungated videos, 55 start at the first unit, five at the second, and one at the third. In the third-unit case, a Seedance~2.0 output uses ``sorry'' as its anchor: it starts at 4.320\,s versus 4.000\,s in the reference, giving $+320$\,ms. This single-anchor exception skips the first two words and has a text-match score of 72.73\%, below the usual 75\% criterion. Later-turn timing is excluded from this opening statistic.

Designated-track SyncNet runs at 25\,fps with 40\,ms lag steps. It requires at least 12 common windows and 0.6 coverage; a peak confidence below one or a boundary optimum leaves the lag unavailable. Expanded-search evidence is retained when inconclusive. The scope comprises longer utterances with established speaker binding. Eight off-screen turns are inapplicable, and 16 source-defined terminal short replies in 11 videos are deferred. Among the 80 in-scope turns after the dialogue gate, 73 have a lag estimate, including one partial sentence window; seven remain unavailable (five weak peaks and two boundary optima), affecting three cases and six videos. Fifty-five videos have values for every in-scope turn, including one with a partial-window flag; five have only some turns and one has none. Distances and peak confidence remain available alongside lag estimates in the evidence records.

\input{tables/condition_coverage.tex}
\input{tables/lip_available.tex}
\FloatBarrier
\input{tables/condition_subject_zh.tex}

\subsection{Interpretation of the measurement tables}
The common subset is determined separately for each metric and reporting group. Partial-role records enter the coverage summary but are excluded from common-case means; measurements with narrowed temporal support can enter those means with scope flags. Thus, numeric completeness refers to the selected roles and intervals. The tables report fidelity on that support, with independent condition-level validation discussed in Appendix~\ref{sec:limitations}.

\input{tables/condition_ranking_zh.tex}

\subsection{Task and subject-setting profiles}
\begin{figure}[!t]
\centering
\includegraphics[width=\linewidth]{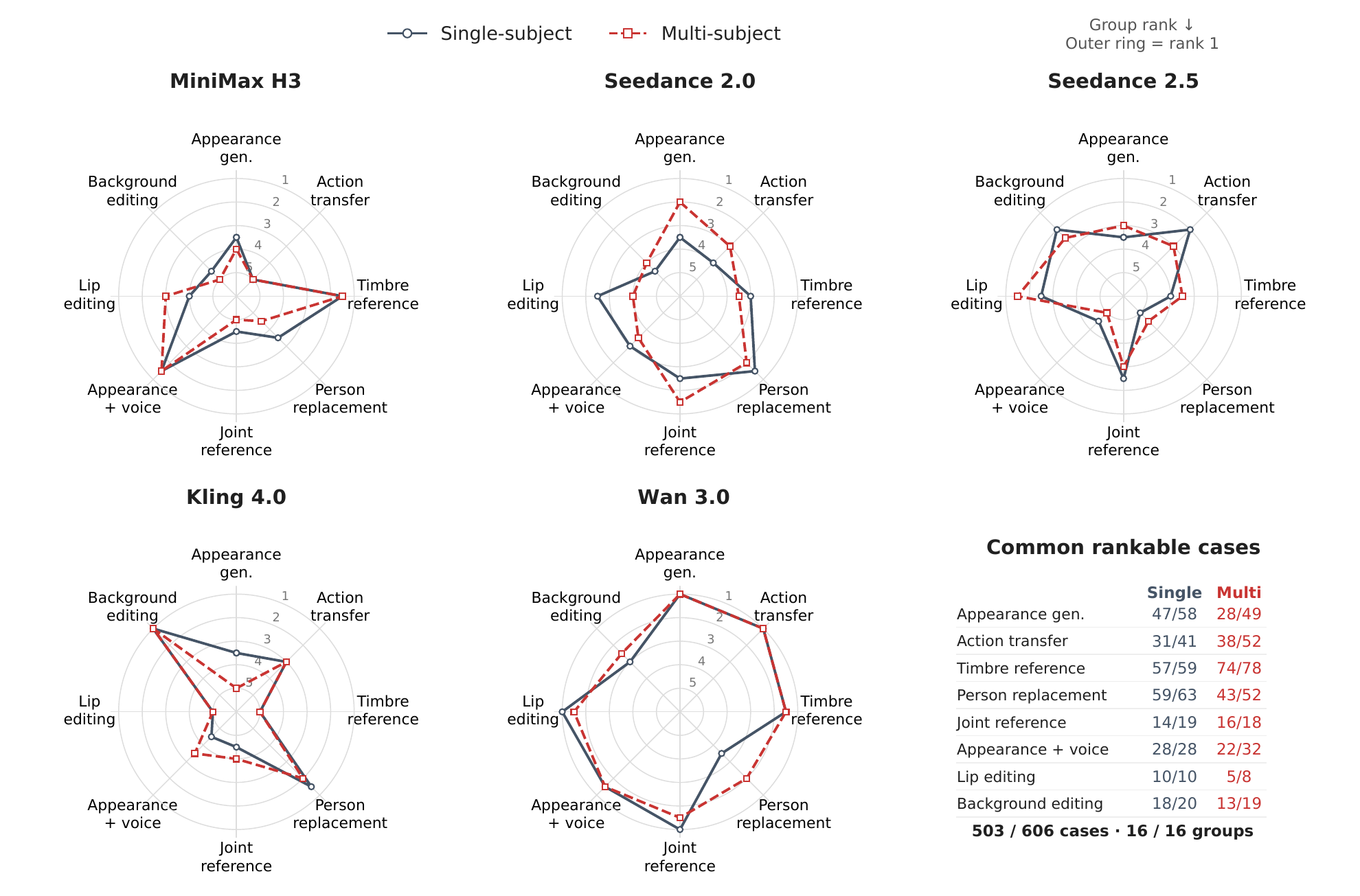}
\caption{Task-condition rank profiles for single- and multi-subject cases. Each model panel plots the tied group positions from Table~\ref{tab:condition-rank-groups}; the outer ring denotes rank 1 and lower ranks are better. The coverage panel reports cases with ranks for all five models / all cases in each group, totaling 503/606 across 16 groups. These are relative task-condition ranks, distinct from human rating scores.}
\label{fig:condition-task-profiles}
\end{figure}

Figure~\ref{fig:condition-task-profiles} visualizes the group positions in Table~\ref{tab:condition-rank-groups}, separating single- and multi-subject cases. Wan 3.0 ranks first in appearance generation and action transfer in both settings. MiniMax H3 and Wan 3.0 share the best positions for timbre-reference generation and appearance-and-voice editing, while Kling 4.0 ranks first for background editing in both settings. Person replacement yields different tied tiers: Seedance~2.0 and Kling 4.0 share the best single-subject position, joined by Wan 3.0 in the multi-subject group. These positions use the fixed within-group tie rule and each group's common rankable cases.

\section{Independent Checklist Evaluation}
\label{sec:checklist-details}

\subsection{Additional MLLM checklist results}
\label{sec:qwen-checklist-results}

Qwen3.8-27B-FP8~\citep{qwen2026qwen38_27b_fp8} evaluates one generated candidate per request using its task caption, role-bound references, and applicable checklist. Ordered visual panels are sampled at 8\,fps; Qwen3-Omni-30B-A3B~\citep{xu2025qwen3omnitechnicalreport} supplies fresh auditory observations and reference comparisons for audio-applicable cases. Caption-blind Qwen3-ASR transcripts support dialogue checking.

We additionally evaluate Qwen3.8-Flash-Next-FP8~\citep{qwen2026flashnext_fp8} using the same captions, checklists, visual inputs, and fixed Qwen3-Omni/ASR evidence. Its visual judgments use fresh single-candidate contexts without the 27B judgments. Appendix Table~\ref{tab:qwen-flashnext} reports this evaluator separately, using the same aggregation rules on its own common scored subset.
Table~\ref{tab:qwen-main} reports task-conditioned checklist judgments for all five models. Across the seven families with comparable task scores, Kling 4.0 leads performance transfer, person replacement, and background editing; Seedance~2.5 leads voice-reference and joint-reference generation; Wan 3.0 leads appearance generation and appearance--voice editing. The rank-based summary places Kling 4.0 first (1.893), followed by Seedance~2.5 (2.893) and Seedance~2.0 (2.929). Kling 4.0 also leads the weighted mean score (2.924), followed by Seedance~2.5 (2.830); both summaries give the same ordering of the five models. Both summaries use 578 common cases across 14 task--subject groups; the Seedance systems are close under this descriptive ranking.

\begin{table}[!t]
\centering
\caption{Independent checklist task scores from Qwen3.8-27B-FP8~\citep{qwen2026qwen38_27b_fp8} with Qwen3-Omni~\citep{xu2025qwen3omnitechnicalreport} auditory evidence (0--4, higher is better). $C/N$ gives cases scored for all five models / task cases. Weighted mean score averages task-score means equally over the 14 groups (higher is better). Overall mean rank (lower is better) averages positions across 14 task--subject groups after within-case ranking. The table covers seven task families with comparable task scores. Bold and underline mark the best and second-best displayed values, including ties.}
\label{tab:qwen-main}
\fontsize{8.5}{10}\selectfont
\setlength{\tabcolsep}{3pt}
\renewcommand{\arraystretch}{1.10}
\begin{tabular*}{\linewidth}{@{\extracolsep{\fill}}lrrrrrr@{}}
\arrayrulecolor{jingdongred}\toprule\arrayrulecolor{black}
\paperth{Task} & \paperth{$C/N$} & \paperth{\shortstack{MiniMax\\H3}} & \paperth{\shortstack{Seedance\\2.0}} & \paperth{\shortstack{Seedance\\2.5}} & \paperth{Kling 4.0} & \paperth{Wan 3.0} \\
\midrule
Appearance & 107/107 & 3.037 & \underline{3.196} & 3.056 & 3.187 & \textbf{3.271} \\
Performance & 93/93 & 2.097 & \underline{2.914} & \underline{2.914} & \textbf{2.957} & 2.828 \\
Voice reference & 134/137 & 3.545 & 3.552 & \textbf{3.701} & \underline{3.687} & 3.366 \\
Person replacement & 115/115 & 1.852 & \underline{2.017} & 1.835 & \textbf{2.061} & 1.809 \\
Joint reference & 34/37 & 2.176 & 2.824 & \textbf{2.912} & \underline{2.853} & 2.618 \\
Appearance + voice & 56/60 & \underline{2.875} & \underline{2.875} & 2.679 & 2.857 & \textbf{2.893} \\
Background & 39/39 & 1.308 & 2.154 & 2.667 & \textbf{2.795} & \underline{2.692} \\
\midrule
\paperth{Weighted mean score $\uparrow$} & 578/606 & 2.420 & 2.797 & \underline{2.830} & \textbf{2.924} & 2.792 \\
\paperth{Overall mean rank $\downarrow$} & 578/606 & 4.143 & 2.929 & \underline{2.893} & \textbf{1.893} & 3.143 \\
\bottomrule
\end{tabular*}
\end{table}

The checklist and metric-based rankings emphasize different evidence. Wan 3.0 leads the condition-measurement ranking in Table~\ref{tab:condition-main}, whereas Kling 4.0 ranked first under the Qwen3.8-27B-FP8~\citep{qwen2026qwen38_27b_fp8} checklist (Table~\ref{tab:qwen-main}). The latter assesses the applicable task requirements jointly from sampled visual and auditory evidence, while the former aggregates condition-specific measurements. The two results therefore describe complementary assessments on their respective common-case sets. For lip editing, 29 of 90 candidates trigger dialogue failure and the other 61 retain uncertain task completion because precise synchronization is outside this checklist's evidence scope; neither subset provides a complete task-level comparison.

\paragraph{Flash-Next checklist assessment.}
Table~\ref{tab:qwen-flashnext} reports task-conditioned checklist judgments from Qwen3.8-Flash-Next-FP8. Seedance~2.5 has the highest mean score for voice-reference generation and background editing, while Seedance~2.0 leads person replacement. Kling 4.0 and Wan 3.0 tie on appearance generation, and both Seedance systems tie with Kling 4.0 on performance transfer. Kling 4.0 leads both overall summaries: a weighted mean score of 3.798 and a mean rank of 2.357. Wan 3.0 is second by weighted score (3.724), whereas Seedance~2.0 is second by mean rank (2.750), reflecting the distinction between averaging score magnitudes and group positions.

The summaries cover 461 common cases across 14 task--subject groups. Joint-reference generation contributes only two common cases, one per subject-count group; appearance--voice editing contributes 14, including one single-person case. Equal group weighting gives these small subsets the same weight as larger groups, so the overall results should be read alongside the reported coverage. Appendix~\ref{sec:flashnext-details} provides the group breakdown.

\begin{table}[!t]
\centering
\caption{Independent checklist task scores from Qwen3.8-Flash-Next-FP8~\citep{qwen2026flashnext_fp8} with Qwen3-Omni~\citep{xu2025qwen3omnitechnicalreport} auditory evidence (0--4, higher is better). $C/N$ gives cases scored for all five models / task cases. Weighted mean score averages task-score means equally over the 14 groups (higher is better). Overall mean rank (lower is better) averages positions across 14 task--subject groups after within-case ranking. The table covers seven task families with comparable task scores. Bold and underline mark the best and second-best displayed values, including ties.}
\label{tab:qwen-flashnext}
\fontsize{8.5}{10}\selectfont
\setlength{\tabcolsep}{3pt}
\renewcommand{\arraystretch}{1.10}
\begin{tabular*}{\linewidth}{@{\extracolsep{\fill}}lrrrrrr@{}}
\arrayrulecolor{jingdongred}\toprule\arrayrulecolor{black}
\paperth{Task} & \paperth{$C/N$} & \paperth{\shortstack{MiniMax\\H3}} & \paperth{\shortstack{Seedance\\2.0}} & \paperth{\shortstack{Seedance\\2.5}} & \paperth{Kling 4.0} & \paperth{Wan 3.0} \\
\midrule
Appearance & 105/107 & 3.857 & 3.905 & \underline{3.914} & \textbf{3.924} & \textbf{3.924} \\
Performance & 93/93 & 2.484 & \textbf{3.849} & \textbf{3.849} & \textbf{3.849} & \underline{3.720} \\
Voice reference & 102/137 & 3.863 & 3.882 & \textbf{3.961} & \underline{3.922} & 3.804 \\
Person replacement & 111/115 & 3.288 & \textbf{3.577} & 3.252 & \underline{3.568} & 3.450 \\
Joint reference & 2/37 & \underline{3.500} & \textbf{4.000} & \underline{3.500} & \textbf{4.000} & \textbf{4.000} \\
Appearance + voice & 14/60 & \underline{3.000} & \textbf{3.286} & 2.857 & \textbf{3.286} & \textbf{3.286} \\
Background & 34/39 & 1.794 & 3.059 & \textbf{3.794} & \underline{3.706} & 3.559 \\
\midrule
\paperth{Weighted mean score $\uparrow$} & 461/606 & 2.978 & 3.701 & 3.602 & \textbf{3.798} & \underline{3.724} \\
\paperth{Overall mean rank $\downarrow$} & 461/606 & 4.036 & \underline{2.750} & 3.071 & \textbf{2.357} & 2.786 \\
\bottomrule
\end{tabular*}
\end{table}

Both evaluators place Kling 4.0 first in their overall summaries. Their common-case coverage differs: 578 cases for the 27B judge and 461 for Flash-Next. Since evaluator and cohort both change, the score differences do not isolate a judge-size effect.

\FloatBarrier

\subsection{Protocol and aggregation}

The checklist evaluation uses fresh single-candidate contexts for all 3,030 outputs. The 0--4 anchors are: 0, an observable requirement is absent or contradicted; 1, only a small part is satisfied with severe deviation; 2, substantial partial satisfaction with a material deviation; 3, mostly satisfied with a minor local deviation; and 4, all observable applicable requirements are satisfied. Appearance-reference generation uses visual conditions, while voice-reference generation uses dialogue, speaker assignment, and timbre conditions. Editing checklists specify both changed and protected attributes. Condition scores and the task score are recorded separately.

For the MLLM evaluation using Qwen3.8-27B-FP8 in Table~\ref{tab:qwen-main}, Qwen3-Omni first listens without a target script, then compares a candidate with its labeled audio references and task requirements. Qwen3-ASR provides a separate caption-blind transcript. WER strictly above 0.50 gates timbre comparison and lip-editing timing/articulation; an unknown gate leaves timbre uncertain. In lip editing, a triggered gate records task completion and dialogue content as zero, while downstream conditions remain gated. Otherwise, opening alignment and precise articulation are deferred and task completion remains uncertain. These states are distinct from interface failures. All 3,030 checklist responses passed output validation; task completion is scored for 2,950 candidates and uncertain for 80, including 61 lip-editing candidates. The remaining 19 uncertain candidates exclude ten cases from the common subset of the other seven tasks, leaving 578 common cases.

For the supplementary MLLM evaluations, the score-based and rank-based summaries use each evaluator's common cases and eligible groups. Let $\bar{s}_{gm}$ be the mean task score for model $m$ over the common cases in group $g$. Its weighted mean score is
\begin{equation}
S_m=\sum_{g\in\mathcal G}\frac{1}{14}\bar{s}_{gm},\qquad |\mathcal G|=14.
\end{equation}
Thus each task--subject group has equal total weight despite different sample counts. This summary averages task scores on their original 0--4 scale; the rank-based summary below instead averages group positions.

For a common case with task scores $s_{im}$, the tied sample rank is
\begin{equation}
r_{im}=1+\sum_{m'\ne m}\mathbf{1}[s_{im'}>s_{im}]+\tfrac12\sum_{m'\ne m}\mathbf{1}[s_{im'}=s_{im}].
\end{equation}
Within each task--subject group $g$, we average $r_{im}$ over common cases. Sorting these means in ascending order, each tie group extends from its smallest mean to subsequent means within 0.20 of that value. Its models receive the mean of the occupied positions. Overall mean rank is the equal average of these group positions across 14 groups. The tolerance is descriptive; it is applied to average sample ranks, not to 0--4 scores, and is not applied again to the overall mean. For the Qwen3.8-27B results, setting the group tolerance to 0, 0.10, 0.20, or 0.30 retains Kling 4.0 first and MiniMax H3 last, while the intermediate order varies.

\input{tables/qwen_checklist_groups.tex}

\subsection{Flash-Next checklist coverage}
\label{sec:flashnext-details}
The Flash-Next evaluation holds the captions, checklists, visual inputs, and Qwen3-Omni/ASR evidence fixed relative to the 27B evaluation, while collecting new visual judgments. All 3,030 responses passed output validation: task completion is scored for 2,635 candidates and uncertain for 395, including 61 lip-editing candidates. The other 334 uncertain candidates exclude 127 cases from the common subset of the seven remaining tasks, leaving 461 common cases. Some uncertain judgments in combination tasks cite unavailable precise synchronization evidence; their original uncertain states are retained. The same score and rank aggregation rules above are applied to the 14 eligible task--subject groups. Joint-reference generation has one common case in each subject-count group; appearance--voice editing has one single-person and 13 multi-person common cases. These small groups retain the same weight as larger groups. Across group tolerances of 0, 0.10, 0.20, and 0.30, Kling 4.0 remains first and MiniMax H3 last, while the intermediate order varies.

\input{tables/qwen_flashnext_groups.tex}

\section{Supplementary Pairwise Evaluation}
\label{sec:automatic-details}
These separately collected pairwise experiments are distinct from the independent checklist evaluation in Table~\ref{tab:qwen-main}; their judgments are not used to construct checklist inputs or scores.
We implement an evaluation protocol inspired by ORAV~\citep{hua2026orav} using our own prompts. The 27B/Omni evaluator compares task completion, visual quality, audio quality, and overall preference, pooling valid outcomes over opponents and cases. Qwen3.8-Flash-Next-FP8~\citep{qiu2026designqwen38nextarchitectureevaluation} supplies a separate task-completion and visual-quality comparison, aggregated by averaging valid opponents within each case and then common cases across systems. The protocols use separate prompts and response schemas; the 27B/Omni rubric scores retain their 0--4 scale. Table~\ref{tab:judge-coverage} reports valid pairs, merged ties, order conflicts, and uncertain pairs.

Each case's five outputs form ten unordered model pairs, each presented in both orders. Conflicting valid order-specific judgments are merged into a tie and flagged; uncertain or failed comparisons are excluded. Wins, ties, and losses contribute 1, 0.5, and 0 to preference scores. Flash-Next confidence intervals use 1,000 task-stratified paired case-ID bootstrap draws.

\begin{table}[!t]
\caption{Automatic-comparison coverage and order consistency by evaluator and dimension. Counts refer to unordered model pairs; Tie includes conflict-merged ties, and U denotes uncertain pairs. Percentages use valid pairs.}
\label{tab:judge-coverage}
\centering

\footnotesize
\setlength{\tabcolsep}{4pt}
\begin{tabular}{p{0.22\linewidth}p{0.22\linewidth}rrrr}
\arrayrulecolor{jingdongred}\toprule\arrayrulecolor{black}
 \paperth{\centering\arraybackslash Version} & \paperth{\centering\arraybackslash Dimension} & \cth{Valid} & \cth{Tie \%} & \cth{Conflict \%} & \cth{U} \\
\midrule
27B/Omni (pooled) & Task completion & 5,658 & 46.540 & 37.500 & 402 \\
\paperrowsep
27B/Omni (pooled) & Visual quality & 6,015 & 92.020 & 19.240 & 45 \\
\paperrowsep
27B/Omni (pooled) & Audio quality & 5,828 & 91.940 & 10.500 & 232 \\
\paperrowsep
27B/Omni (pooled) & Overall preference & 5,674 & 43.830 & 36.320 & 386 \\
\paperrowsep
Flash-Next (case mean) & Task completion & 5,880 & 46.120 & 39.440 & 180 \\
\paperrowsep
Flash-Next (case mean) & Visual quality & 6,060 & 37.130 & 37.130 & 0 \\
\bottomrule
\end{tabular}
\end{table}

\textbf{ Order consistency.}
Presentation order materially affects the automatic comparisons. For Flash-Next, 2,319 of 5,880 valid task-completion pairs conflict across the two orders, or 39.44\%; 2,250 of 6,060 visual-quality pairs conflict, or 37.13\%. All merged visual ties arise from these conflicts. Task-completion coverage comprises 588 common cases; the eighteen supplied-recording lip-editing cases contribute 180 uncertain pairs. Visual-quality judgments cover all 606 cases.

These observations make order consistency part of the reported measurement. We retain conflicts separately from agreed ties and keep evaluator configurations distinct. A balanced fixed-input audit also changes the resulting preferences when prompt, response schema, evidence layout, and output budget are revised together. Appendix~\ref{sec:judge-audit} gives the comparison counts and intervals.

\section{Fixed-Input Visual-Evaluator Audit}
\label{sec:judge-audit}
We compare two Qwen3.8-Flash-Next-FP8 configurations on eighty fixed cases, ten per task family with five single-subject and five multi-subject cases each. Both configurations receive identical no-padding candidate image blocks. Configuration B differs from configuration A in its prompt, response schema, evidence-first response layout, and output budget (2,200 versus 1,400 tokens).

\begin{table}[!t]
\caption{Fixed-input visual judging of Seedance 2.0~\citep{seedance2026seedance20advancingvideo} and 2.5 on 80 cases. Both configurations use the same candidate images and yield valid comparisons throughout. Conflicts are included in Ties; preference scores assign half credit to ties.}
\centering

\footnotesize
\setlength{\tabcolsep}{4pt}
\begin{tabular}{lrrrrrr}
\arrayrulecolor{jingdongred}\toprule\arrayrulecolor{black}

\cth{Configuration} & \cth{Wins 2.0} & \cth{Wins 2.5} & \cth{Ties} & \cth{Conflicts} & \cth{Pref.~2.0} & \cth{Pref.~2.5} \\
\midrule
Configuration A & 47 & 5 & 28 & 28 & 76.250 & 23.750 \\
\paperrowsep
Configuration B & 30 & 8 & 42 & 40 & 63.750 & 36.250 \\
\bottomrule
\end{tabular}
\end{table}

Seedance~2.0's preference score changes from 76.25 to 63.75, while Seedance~2.5's changes from 23.75 to 36.25. The order-conflict rate increases from 35\% to 50\%, a difference of 15 percentage points with a paired-bootstrap 95\% interval of [2.5, 27.5]. Forty of configuration B's 42 merged ties arise from order conflicts; two are agreed ties. Its forty conflicts comprise 28 opposite-model winners and twelve winner-versus-tie cases; 22 were stable under configuration A. These annotations describe inconsistencies between order-specific responses.

The audit intervals use 5,000 paired case-ID bootstrap draws stratified by task family. A separate padding control holds the prompt, response schema, and decoding fixed. Removing padding changes Seedance~2.0's preference score by $-1.875$ percentage points, with 95\% interval [$-6.25$, 1.875]. The padding control and the configuration comparison test different changes.

\section{Diagnostic Rubric}
\label{sec:diagnostic-rubric}
This section reports the separate 39-criterion pairwise diagnostic rubric, not the task-specific checklist used in Table~\ref{tab:qwen-main}.
The thirty-nine criteria distinguish requested attributes, actor assignment, source retention, and output quality. Table~\ref{tab:criterion-facets} groups these components. Of the 65,851 scored output-by-criterion assessments, 47,797 concern requested content, reference control, or source preservation, and 18,054 assess perceptual quality. The 13,267 distinct case-level checks pair each case with a criterion scored for at least one system, counted once across systems and repeated judgments.

Repeated opponent/order ratings are averaged within each case, system, and criterion before scored cases are aggregated. Uncertain, unavailable, and inapplicable observations retain separate states.

\begin{table}[!t]
\caption{The 39 automatic diagnostic criteria, organized by control and quality requirements. Count denotes the number of criteria in each group.}
\label{tab:criterion-facets}
\centering

\footnotesize
\setlength{\tabcolsep}{4pt}
\begin{tabular}{p{0.28\linewidth}rp{0.56\linewidth}}
\arrayrulecolor{jingdongred}\toprule\arrayrulecolor{black}
 \paperth{\centering\arraybackslash Criterion group} & \cth{Count} & \paperth{\centering\arraybackslash Component requirements} \\
\midrule
Text and soundscape & 3 & Scene/camera, pose/expression, and soundscape content \\
\paperrowsep
Appearance and assignment & 6 & Identity, attributes, visual binding, stability, hairstyle, visible build \\
\paperrowsep
Performance and interaction & 4 & Action, sequence, pose/expression/gaze, interaction \\
\paperrowsep
Speech, voice, soundtrack & 7 & Words, turns, timbre, voice stability, speaker binding, soundtrack speech/background \\
\paperrowsep
Editing and retention & 5 & Edit target, non-targets, scene/camera, source clothing, source identity \\
\paperrowsep
Background and integration & 7 & Characteristics, structure, adaptation, light, viewpoint, contacts, boundaries \\
\paperrowsep
Perceptual quality & 6 & Visual clarity/anatomy/temporal behavior; audio intelligibility/artifacts/continuity \\
\paperrowsep
Articulation adaptation & 1 & Lip-articulation adaptation \\
\bottomrule
\end{tabular}
\end{table}

Table~\ref{tab:criterion-results} reports all thirty-nine scored criteria. Each row uses the intersection of cases with numeric scores for all five models; repeated opponent/order scores are averaged within each case and system before cases receive equal weight. Unlike the two multi-subject assignment rows in Table~\ref{tab:control-diagnostics}, these rows include both subject groups wherever the criterion is scored.

\begin{table}[!p]
\caption{MLLM diagnostic scores for all 39 criteria (0--4, higher is better). Cases gives the subset with scores for all five models on each criterion. Bold and underline mark the best and second-best displayed means, including ties.}
\label{tab:criterion-results}
\centering

\footnotesize
\setlength{\tabcolsep}{4pt}
\renewcommand{\arraystretch}{1.03}
\begin{tabular}{@{}p{0.40\linewidth}rrrrrr@{}}
\arrayrulecolor{jingdongred}\toprule\arrayrulecolor{black}
 \paperth{\centering\arraybackslash Criterion} & \cth{Cases} & \cth{\shortstack{MiniMax\\H3}} & \cth{\shortstack{Seedance\\2.0}} & \cth{\shortstack{Seedance\\2.5}} & \cth{Kling 4.0} & \cth{Wan 3.0} \\
\midrule
\rowcolor{tablesection}
\multicolumn{7}{@{}l}{\textbf{Text and soundscape}} \\
Scene and camera adherence & 524 & 3.670 & \underline{3.770} & 3.760 & \textbf{3.840} & 3.690 \\
\paperrowsep
Pose and expression adherence & 519 & 3.720 & \underline{3.760} & \underline{3.760} & \textbf{3.840} & 3.750 \\
\paperrowsep
Soundscape adherence & 437 & 3.020 & \underline{3.180} & 3.130 & 3.110 & \textbf{3.250} \\
\midrule
\rowcolor{tablesection}
\multicolumn{7}{@{}l}{\textbf{Appearance and assignment}} \\
Appearance identity & 606 & 3.320 & 3.770 & 3.630 & \textbf{3.840} & \underline{3.800} \\
\paperrowsep
Appearance attributes & 593 & 3.390 & \underline{3.760} & 3.650 & \textbf{3.860} & \underline{3.760} \\
\paperrowsep
Visual-reference assignment & 478 & 3.620 & 3.920 & 3.860 & \textbf{3.950} & \underline{3.930} \\
\paperrowsep
Identity stability & 606 & 3.930 & \underline{3.970} & 3.950 & \textbf{3.980} & \textbf{3.980} \\
\paperrowsep
Referenced hairstyle & 4 & 3.030 & 2.160 & \underline{3.470} & 3.340 & \textbf{3.780} \\
\paperrowsep
Referenced visible build & 4 & 2.780 & 1.930 & \underline{2.990} & 2.820 & \textbf{3.200} \\
\midrule
\rowcolor{tablesection}
\multicolumn{7}{@{}l}{\textbf{Performance and interaction}} \\
Action completion & 603 & 3.670 & 3.720 & 3.740 & \textbf{3.820} & \underline{3.780} \\
\paperrowsep
Performance sequence & 357 & 3.640 & 3.750 & 3.770 & \textbf{3.840} & \underline{3.780} \\
\paperrowsep
Pose, expression, and gaze & 361 & 3.690 & 3.810 & \underline{3.840} & \textbf{3.900} & 3.820 \\
\paperrowsep
Person interaction & 361 & 3.790 & \underline{3.890} & 3.880 & \textbf{3.920} & \underline{3.890} \\
\midrule
\rowcolor{tablesection}
\multicolumn{7}{@{}l}{\textbf{Speech, voice, and soundtrack}} \\
Dialogue content & 563 & 3.500 & 3.620 & \textbf{3.690} & \underline{3.650} & 3.590 \\
\paperrowsep
Dialogue turns & 563 & 3.710 & 3.800 & \textbf{3.850} & 3.750 & \underline{3.830} \\
\paperrowsep
Timbre similarity & 234 & \underline{3.340} & 3.330 & 3.240 & 3.150 & \textbf{3.360} \\
\paperrowsep
Voice stability & 560 & 3.970 & \textbf{3.990} & \underline{3.980} & \textbf{3.990} & \underline{3.980} \\
\paperrowsep
Speaker assignment & 561 & 3.480 & 3.650 & \underline{3.700} & 3.600 & \textbf{3.710} \\
\paperrowsep
Soundtrack speech preservation & 172 & 3.090 & 3.590 & \underline{3.680} & 3.370 & \textbf{3.700} \\
\paperrowsep
Soundtrack background preservation & 172 & 3.340 & 3.770 & \textbf{3.840} & 3.470 & \underline{3.800} \\
\midrule
\rowcolor{tablesection}
\multicolumn{7}{@{}l}{\textbf{Editing and retention}} \\
Requested edit & 232 & 3.370 & 3.440 & 3.250 & \textbf{3.760} & \underline{3.660} \\
\paperrowsep
Non-target preservation & 227 & 3.570 & \underline{3.780} & \textbf{3.810} & \textbf{3.810} & \underline{3.780} \\
\paperrowsep
Scene and camera preservation & 323 & 3.780 & 3.780 & \underline{3.820} & \textbf{3.910} & 3.810 \\
\paperrowsep
Source clothing preservation & 80 & 3.390 & \textbf{3.810} & \underline{3.730} & 3.480 & 3.480 \\
\paperrowsep
Source identity preservation & 57 & 3.640 & 3.860 & \textbf{3.940} & 3.910 & \underline{3.930} \\
\midrule
\rowcolor{tablesection}
\multicolumn{7}{@{}l}{\textbf{Background and integration}} \\
Background characteristics & 39 & 2.480 & 2.800 & 3.450 & \textbf{3.930} & \underline{3.690} \\
\paperrowsep
Background structure & 19 & 2.640 & 2.680 & 2.990 & \textbf{3.800} & \underline{3.520} \\
\paperrowsep
Background adaptation & 20 & 2.090 & 2.640 & \underline{3.610} & \textbf{3.810} & 3.560 \\
\paperrowsep
Lighting integration & 39 & 2.630 & 2.870 & 3.290 & \textbf{3.440} & \underline{3.410} \\
\paperrowsep
Viewpoint integration & 39 & 2.520 & 3.030 & 3.580 & \textbf{3.730} & \underline{3.710} \\
\paperrowsep
Foreground contact preservation & 20 & 3.450 & 3.660 & \textbf{3.900} & 3.670 & \underline{3.740} \\
\paperrowsep
Edit-boundary integration & 62 & 2.980 & 3.290 & 3.630 & \textbf{3.780} & \underline{3.710} \\
\midrule
\rowcolor{tablesection}
\multicolumn{7}{@{}l}{\textbf{Perceptual quality}} \\
Visual clarity & 606 & 3.820 & \underline{3.920} & 3.800 & \textbf{3.940} & 3.860 \\
\paperrowsep
Anatomical plausibility & 606 & 3.970 & \underline{3.980} & 3.960 & \textbf{3.990} & \underline{3.980} \\
\paperrowsep
Temporal visual quality & 606 & 3.960 & \underline{3.980} & 3.960 & \textbf{3.990} & \underline{3.980} \\
\paperrowsep
Audio intelligibility & 567 & 3.870 & \underline{3.970} & \underline{3.970} & \textbf{3.980} & \underline{3.970} \\
\paperrowsep
Freedom from audio artifacts & 602 & 3.920 & 3.950 & 3.950 & \textbf{3.970} & \underline{3.960} \\
\paperrowsep
Audio continuity & 602 & \underline{3.980} & \underline{3.980} & \underline{3.980} & \textbf{3.990} & \textbf{3.990} \\
\midrule
\rowcolor{tablesection}
\multicolumn{7}{@{}l}{\textbf{Articulation adaptation}} \\
Lip-articulation adaptation & 18 & 2.910 & \textbf{3.090} & \underline{3.030} & 2.610 & \textbf{3.090} \\
\bottomrule
\end{tabular}
\end{table}

\subsection{Selected control criteria and paired dialogue analysis}
The automatic rubric assigns its highest mean scores to different systems across control requirements. Table~\ref{tab:control-diagnostics} reports task-aware ratings for actor assignment, performance sequence, and selective preservation on criterion-specific common cases. Wan 3.0 receives the highest mean visual-reference assignment score in multi-person scenes, 3.95, whereas Seedance~2.5 receives the highest mean speaker-assignment score, 3.66. Kling 4.0 has the highest mean performance-sequence score, 3.84, while Seedance~2.0 has the highest mean source-clothing preservation score, 3.81. These automatic judgments separate the rubric questions about reference assignment, performance, and source preservation. They are interpreted together with the pairwise evaluator diagnostics in Table~\ref{tab:judge-coverage}. Table~\ref{tab:criterion-results} reports all thirty-nine criteria.

\begin{table}[!t]
\caption{Selected MLLM control scores on subsets with scores for all five models (0--4, higher is better). Multi restricts evaluation to multi-subject cases; Both includes both subject settings. Bold and underline mark the best and second-best displayed means, including ties.}
\label{tab:control-diagnostics}
\centering

\footnotesize
\setlength{\tabcolsep}{3pt}
\renewcommand{\arraystretch}{1.08}
\begin{tabular}{@{}p{0.29\linewidth}ccrrrrr@{}}
\arrayrulecolor{jingdongred}\toprule\arrayrulecolor{black}
 \paperth{\centering\arraybackslash Criterion} & \cth{Subjects} & \cth{Cases} & \cth{\shortstack{MiniMax\\H3}} & \cth{\shortstack{Seedance\\2.0}} & \cth{\shortstack{Seedance\\2.5}} & \cth{Kling 4.0} & \cth{Wan 3.0} \\
\midrule
Visual-reference assignment & Multi & 308 & 3.630 & 3.910 & 3.870 & \underline{3.930} & \textbf{3.950} \\
\paperrowsep
Speaker assignment & Multi & 300 & 3.420 & 3.610 & \textbf{3.660} & 3.480 & \underline{3.640} \\
\paperrowsep
Performance sequence & Both & 357 & 3.640 & 3.750 & 3.770 & \textbf{3.840} & \underline{3.780} \\
\paperrowsep
Source clothing preservation & Both & 80 & 3.390 & \textbf{3.810} & \underline{3.730} & 3.480 & 3.480 \\
\paperrowsep
Non-target preservation & Both & 227 & 3.570 & \underline{3.780} & \textbf{3.810} & \textbf{3.810} & \underline{3.780} \\
\bottomrule
\end{tabular}
\end{table}

\paragraph{Changing the script with fixed references.}
The ten dialogue pairs compare different requested words under fixed media inputs and actor assignments. Table~\ref{tab:dialogue-contrast} shows that, for four systems, mean dialogue-content scores change little while mean timbre scores decrease. Wan 3.0 instead shows lower dialogue-content and slightly higher timbre scores. The paired rubric profiles therefore distinguish script adherence from timbre similarity under the same reference conditions. Each condition has one generated output per system and reference set.

\begin{table}[!t]
\centering
\caption{MLLM dialogue-content and timbre scores for existing and new scripts on ten paired reference sets (0--4, higher is better). Ordered media inputs and actor assignments are fixed within each pair. Bold and underline mark the best and second-best means in each column, including ties.}
\label{tab:dialogue-contrast}
\small
\begin{tabular}{lrrrr}
\arrayrulecolor{jingdongred}\toprule\arrayrulecolor{black}

 & \multicolumn{2}{c}{Dialogue content} & \multicolumn{2}{c}{Timbre similarity} \\
\cmidrule(lr){2-3}\cmidrule(lr){4-5}
System & Existing & New & Existing & New \\
\midrule
MiniMax H3 & 3.790 & 3.830 & 3.520 & 3.420 \\
\paperrowsep
Seedance 2.0 & 3.770 & 3.830 & \textbf{3.690} & \underline{3.510} \\
\paperrowsep
Seedance 2.5 & \underline{3.880} & \underline{3.920} & \underline{3.670} & 3.450 \\
\paperrowsep
Kling 4.0 & \textbf{3.890} & \textbf{3.940} & 3.590 & 3.490 \\
\paperrowsep
Wan 3.0 & 3.690 & 3.350 & \textbf{3.690} & \textbf{3.710} \\
\bottomrule
\end{tabular}
\end{table}

\clearpage
\section{Attribute and Preservation Diagnostics}
\label{sec:attribute-details}
Person replacement requires different measurements for text-described appearances and image-defined identities. Table~\ref{tab:preservation-diagnostics} separates the three variants using the same metric inventory and common-case aggregation as Table~\ref{tab:condition-main}. The identity margin is inapplicable where no competing reference exists. Text edits report the face, hair, and clothing regions in addition to the whole-person CLIP value. Image edits report the margin between assigned and wrong identity references where competing references exist, alongside source-garment retention. Table~\ref{tab:condition-coverage} reports the available support for hair and torso measurements.

\input{tables/preservation_variants_zh.tex}

For the image-editing variants with source-recording reuse, waveform and spectral cosines describe the compared recording interval. They remain separate from ECAPA voice-reference similarity, which concerns a speaker's voice characteristics even when the requested words differ. Dialogue CER/WER and timbre measurements for voice-conditioned tasks are reported with their task and subject-count masks in Appendix~\ref{sec:condition-details}. This preserves the distinction between generating new words in a specified voice and retaining a supplied recording.

\section{General Automatic Quality Measurements}
\label{sec:quality-details}
The main quality table (Table~\ref{tab:quality-diagnostics}) reports visual, motion, and audio measurements introduced in Section~\ref{sec:evaluation}. Each row uses the same 606 cases across the five models, with cases weighted equally. MUSIQ~\citep{ke2021musiq} uses the SPAQ-trained checkpoint; DOVER++~\citep{wu2023dover} supplies separate technical and aesthetic outputs. AMT-S~\citep{li2023amt} is the interpolation backbone used for the smoothness measurement. Audiobox Aesthetics~\citep{tjandra2025aes} supplies the production-quality (PQ) estimate.

Continuous table entries use three decimal places, while counts remain integers. Bold and underline follow the displayed values, including rounding ties; all calculations retain their original precision. Adjacent-frame CLIP/DINO cosine similarities and VBench consistency scores retain their respective implementations. Dynamic degree describes motion quantity and is not ranked.

The table labels the scale or transformation used for each estimator.

\section{Human Assessment Details}
\label{sec:human-details}
Human assessment covers a reviewed subset of the benchmark. We retain the latest record for each case and reviewer and exclude incomplete records. All five models share the same case set for visual, audio, and independently entered overall scores. Each dimension uses a 1--10 scale. Repeated ratings are averaged within a case before cases receive equal weight. The rating interface displays model names.

The 95\% confidence intervals use 3,000 paired case-ID bootstrap draws stratified by task family, preserving system correspondence and the within-case rating averages. Aggregate scores follow the rated subset's task composition.

\begin{table}[!t]
\centering
\caption{Human visual, audio, and independently entered overall ratings on the reviewed subset (1--10, higher is better). Brackets indicate 95\% task-stratified paired-case bootstrap intervals. Model names were visible to reviewers. Bold and underline mark the best and second-best displayed means, including ties.}
\label{tab:human-ratings}
\fontsize{8.5}{10}\selectfont
\setlength{\tabcolsep}{4pt}
\renewcommand{\arraystretch}{1.10}
\begin{tabular*}{\linewidth}{@{\extracolsep{\fill}}lccc@{}}
\arrayrulecolor{jingdongred}\toprule\arrayrulecolor{black}
 \paperth{System} & \paperth{Visual $\uparrow$} & \paperth{Audio $\uparrow$} & \paperth{Overall $\uparrow$} \\
\midrule
MiniMax H3 & $6.750\,[6.500, 7.020]$ & $7.440\,[7.200, 7.680]$ & $6.690\,[6.460, 6.930]$ \\
\paperrowsep
Seedance 2.0 & $7.810\,[7.610, 7.990]$ & $8.230\,[8.020, 8.420]$ & $7.660\,[7.490, 7.830]$ \\
\paperrowsep
Seedance 2.5 & $\underline{8.270}\,[8.110, 8.430]$ & $\mathbf{8.770}\,[8.640, 8.910]$ & $\mathbf{8.240}\,[8.100, 8.390]$ \\
\paperrowsep
Kling 4.0 & $\mathbf{8.610}\,[8.460, 8.770]$ & $7.240\,[7.060, 7.420]$ & $7.700\,[7.550, 7.850]$ \\
\paperrowsep
Wan 3.0 & $8.220\,[8.060, 8.390]$ & $\underline{8.390}\,[8.220, 8.560]$ & $\underline{8.140}\,[7.980, 8.300]$ \\
\bottomrule
\end{tabular*}
\end{table}

Table~\ref{tab:human-families} reports family-level means for visual, audio, and overall ratings.

\begin{table}[!t]
\caption{Human ratings by task family on the reviewed subset (1--10, higher is better). V, A, and O denote visual, audio, and independently entered overall scores. Bold and underline mark the best and second-best displayed means within each family and dimension, including ties.}
\label{tab:human-families}
\centering

\footnotesize
\setlength{\tabcolsep}{1.6pt}
\renewcommand{\arraystretch}{1.12}
\begin{tabular}{@{}>{\raggedright\arraybackslash}m{0.185\linewidth}*{15}{r}@{}}
\arrayrulecolor{jingdongred}\toprule\arrayrulecolor{black}

\centering\arraybackslash \textbf{Task family} & \multicolumn{3}{c}{\textbf{MiniMax H3}} & \multicolumn{3}{c}{\textbf{Seedance~2.0}} & \multicolumn{3}{c}{\textbf{Seedance~2.5}} & \multicolumn{3}{c}{\textbf{Kling 4.0}} & \multicolumn{3}{c}{\textbf{Wan 3.0}} \\
\cmidrule(lr){2-4}\cmidrule(lr){5-7}\cmidrule(lr){8-10}\cmidrule(lr){11-13}\cmidrule(l){14-16}
 & \cth{V} & \cth{A} & \cth{O} & \cth{V} & \cth{A} & \cth{O} & \cth{V} & \cth{A} & \cth{O} & \cth{V} & \cth{A} & \cth{O} & \cth{V} & \cth{A} & \cth{O} \\
\midrule
Appearance generation & 8.510 & 8.340 & 8.300 & 8.550 & \underline{8.670} & \underline{8.420} & \textbf{8.870} & \textbf{9.070} & \textbf{8.790} & \underline{8.650} & 8.380 & 8.380 & 8.250 & 8.370 & 8.210 \\
\paperrowsep
\addlinespace[3pt]
Performance transfer & 3.950 & 5.110 & 4.210 & 7.360 & 7.610 & 7.310 & 7.690 & \textbf{8.310} & 7.850 & \textbf{8.310} & 8.000 & \textbf{8.120} & \underline{7.800} & \underline{8.060} & \underline{7.880} \\
\paperrowsep
\addlinespace[3pt]
Voice-reference generation & 8.640 & 8.480 & 8.160 & 9.040 & \underline{8.800} & \underline{8.720} & \underline{9.160} & \textbf{9.280} & \textbf{8.960} & \textbf{9.200} & 4.440 & 6.560 & 9.000 & 8.360 & 8.480 \\
\paperrowsep
\addlinespace[3pt]
Joint reference generation & 3.800 & \underline{9.200} & 4.400 & 8.100 & \textbf{9.300} & \underline{7.800} & \underline{8.300} & \textbf{9.300} & \textbf{8.000} & \textbf{9.600} & 5.400 & 7.100 & 7.100 & 8.900 & 7.100 \\
\paperrowsep
\addlinespace[3pt]
Appearance editing & 7.430 & \underline{8.870} & 7.480 & 6.960 & 7.830 & 6.650 & 7.650 & 8.740 & 7.570 & \textbf{8.570} & 8.830 & \underline{8.220} & \underline{8.430} & \textbf{9.040} & \textbf{8.520} \\
\paperrowsep
\addlinespace[3pt]
Appearance and voice editing & 7.550 & \textbf{9.270} & \underline{7.730} & 5.910 & \underline{8.090} & 6.000 & 7.820 & 7.640 & 7.450 & \textbf{9.270} & 5.730 & 7.640 & \underline{8.450} & \underline{8.090} & \textbf{8.090} \\
\paperrowsep
\addlinespace[3pt]
Supplied-recording lip editing & 6.800 & 7.600 & 6.300 & \underline{8.500} & 8.400 & 7.900 & \underline{8.500} & \textbf{9.700} & \textbf{8.900} & 8.200 & 1.000 & 3.000 & \textbf{9.000} & \underline{8.800} & \underline{8.500} \\
\paperrowsep
\addlinespace[3pt]
Background editing & 3.130 & 3.500 & 3.080 & 5.580 & 6.840 & 5.790 & 6.740 & \underline{7.790} & 6.890 & \underline{7.740} & 7.550 & \underline{7.530} & \textbf{7.820} & \textbf{8.130} & \textbf{7.890} \\
\midrule
\rowcolor{comavrow}
All families & 6.750 & 7.440 & 6.690 & 7.810 & 8.230 & 7.660 & \underline{8.270} & \textbf{8.770} & \textbf{8.240} & \textbf{8.610} & 7.240 & 7.700 & 8.220 & \underline{8.390} & \underline{8.140} \\
\bottomrule
\end{tabular}
\end{table}

The human rankings combine manually adjusted orderings and automatic orderings derived from the independently entered overall scores. Tied positions receive their midpoint rank. We average ranks within cases and then across cases, using the same complete subset and 3,000-draw stratified paired case bootstrap as the score analysis. Table~\ref{tab:human-ranks} reports these descriptive ranks.

\begin{table}[!t]
\caption{Mean human rank on the reviewed subset (lower is better), with 95\% paired-case bootstrap intervals. Orders combine manual adjustments and orders derived from overall scores. Bold and underline mark the lowest and second-lowest means.}
\label{tab:human-ranks}
\centering

\footnotesize
\begin{tabular}{lc}
\arrayrulecolor{jingdongred}\toprule\arrayrulecolor{black}

\cth{System} & \cth{Mean rank $\downarrow$} \\
\midrule
MiniMax H3 & $3.560\,[3.410, 3.710]$ \\
\paperrowsep
Seedance~2.0 & $3.090\,[2.940, 3.240]$ \\
\paperrowsep
Seedance~2.5 & $\mathbf{2.500}\,[2.360, 2.640]$ \\
\paperrowsep
Kling 4.0 & $3.050\,[2.910, 3.200]$ \\
\paperrowsep
Wan 3.0 & $\underline{2.800}\,[2.640, 2.960]$ \\
\bottomrule
\end{tabular}
\end{table}

Table~\ref{tab:human-families} reports the reviewed task subsets separately. Seedance~2.5 has the highest mean overall human rating for appearance generation, voice-reference generation, joint reference generation, and supplied-recording lip editing. Wan 3.0 leads appearance editing, joint appearance-and-voice editing, and background editing; Kling 4.0 leads performance transfer. These human ratings complement the task-condition profiles in Figure~\ref{fig:condition-task-profiles}.

\FloatBarrier
\section{Limitations and Responsible Use}
\label{sec:limitations}
\textbf{Coverage and generalization.} The benchmark has unequal task sizes, shared source material, and one output per system and case. Common-case condition comparisons can favor easier-to-measure outputs; the broader identity/hair/build variant contains only four cases. Human ratings cover an uneven task mixture with visible model names and a fixed set of ratings. Repeated generations, broader source sampling, and more balanced human coverage would strengthen generalization beyond these evaluated cases.

\textbf{Measurement validity.} Detectors, ASR, alignment, speaker embeddings, and synchronization models provide learned estimates. Manually assisted timing or role confirmation, short speech, occlusion, and partial sentence windows create heterogeneous measurement support. Independent appearance matching and single-frame evidence provide limited information about actor binding and temporal stability; endpoint pose and shared-prefix audio cover selected portions of performance and recording preservation. Lip synchronization covers selected long utterances, and checklist judgments depend on the available sampled evidence. Independent condition-level annotation and held-out calibration are needed to establish how these estimates relate to complete task satisfaction.

\textbf{Source provenance and use.} Shared clips and identities limit case independence; training exposure and actor-disjointness are unresolved. Public release of the original videos and associated code and data is pending company review and approval.

\section{Caption Length and Task-Condition Analyses}
\begin{figure}[!t]
\centering
\includegraphics[width=\linewidth]{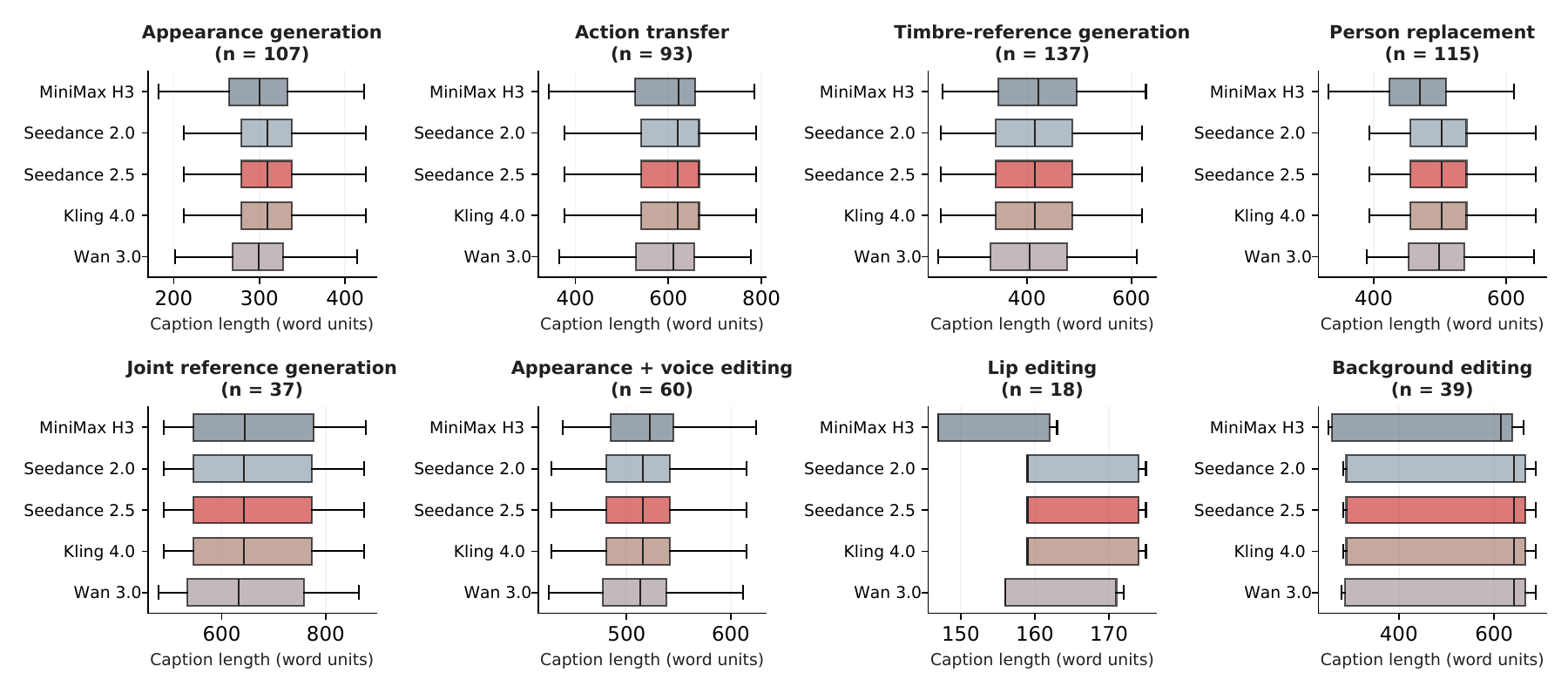}
\caption{Caption lengths by task family and model. Boxes show the median and interquartile range; whiskers reach the most extreme observations within 1.5 interquartile ranges of the box, with outliers omitted. One alphanumeric word or Chinese character counts as one unit.}
\label{fig:prompt-distributions}
\end{figure}

\subsection{Task-condition ranks by subject setting}
\begin{figure}[!t]
\centering
\includegraphics[width=\linewidth]{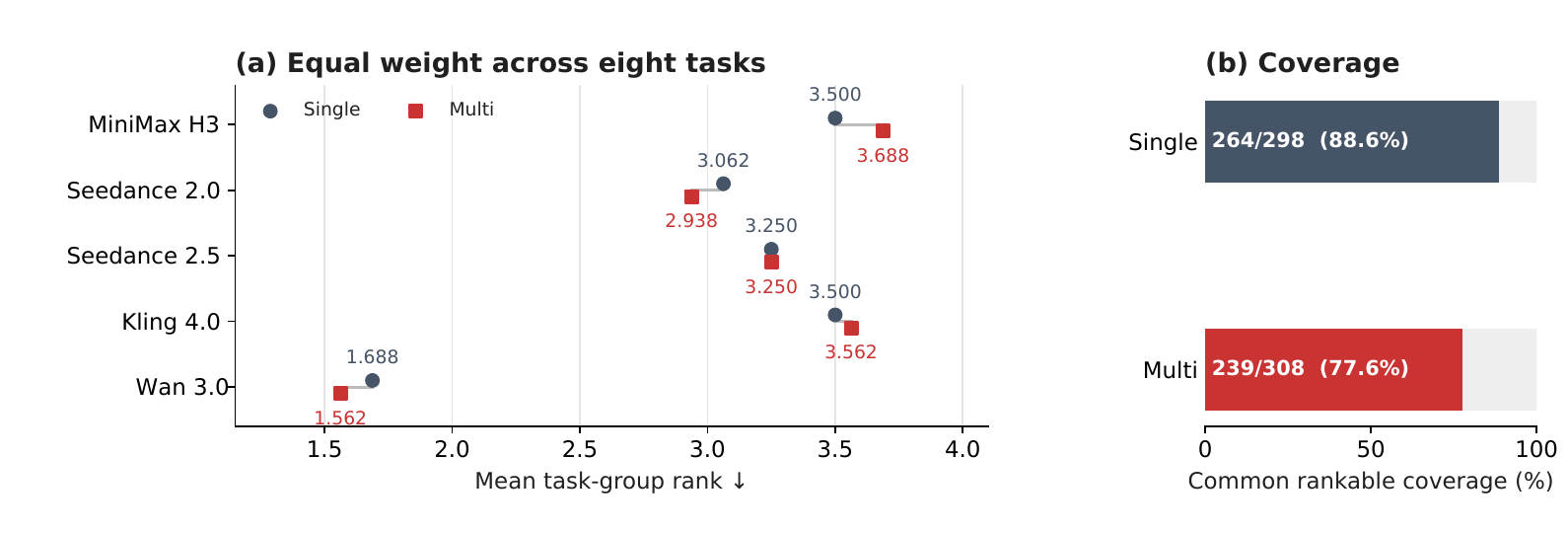}
\caption{Task-condition ranks and common rankable coverage by subject setting. (a) Each point is the equal-weight mean of eight task-group positions; lower is better. Connecting segments pair the same model across settings. (b) Coverage counts cases with determined comparisons among all five models. The two settings contain different cases.}
\label{fig:condition-subject-groups}
\end{figure}

Figure~\ref{fig:condition-subject-groups} averages the eight task-group positions separately for each subject setting, retaining the same equal-task weighting as the overall rank. Common rankable coverage is 264/298 for single-subject cases and 239/308 for multi-subject cases. Wan 3.0 ranks first in both settings, with mean ranks of 1.688 and 1.562, followed by Seedance~2.0 (3.062 and 2.938); Seedance~2.5 has a mean of 3.250 in both. These setting-specific summaries average back to the overall values in Table~\ref{tab:condition-rank-groups}.

\begin{figure}[!t]
\centering
\includegraphics[width=\linewidth]{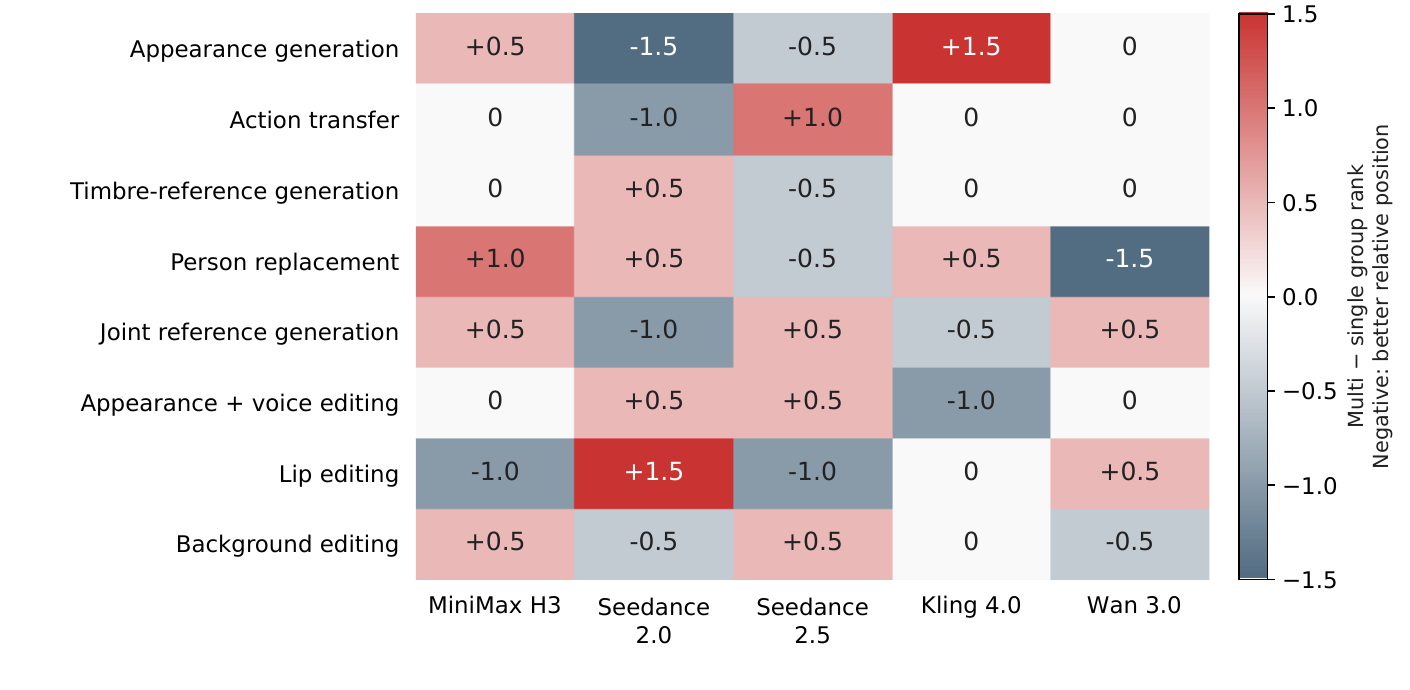}
\caption{Task-wise changes in model position: multi-subject minus single-subject group rank. Negative values indicate a better relative position in the multi-subject group, positive values a worse position, and zero an unchanged position. Cells use the same tied positions and common rankable subsets as Figure~\ref{fig:condition-task-profiles}.}
\label{fig:condition-within-family}
\end{figure}

The task-wise shifts in Figure~\ref{fig:condition-within-family} reveal differences hidden by the aggregate means. Seedance~2.0 moves from group position 3.5 to 2.0 for appearance generation but from 2.5 to 4.0 for lip editing. Kling 4.0 remains first for background editing in both settings, while its appearance-generation position changes from 3.5 to 5.0. These shifts describe relative model ordering across distinct task/subject groups.

\FloatBarrier
\section{Representative Captions and Input Roles}
\label{sec:caption-examples}
The excerpts below retain the original wording of the final task descriptions. They illustrate reference roles, dialogue assignments, and preservation requirements. The two longer editing examples show complete prompts for the same case, adapted to MiniMax H3 and Seedance 2.5. In these quoted prompts, ``silent source'' denotes a visual-only video input; it does not require the original file or generated output to be silent. Audio requirements follow the specified soundtrack or timbre references.

\subsection{Identity replacement with source-clothing retention -- MiniMax H3}
\noindent Example A (MiniMax H3).
\begingroup\small\raggedright
subject\_definitions:
<Subject 1> is the replacement for the lone foreground person in <Video 1>. Facial identity, hairstyle and visible body build come from the bound person in <Picture 1>: the primary named person, centred in the photograph. Ignore all other people in that photograph. Clothing, headwear, accessories, pose and performance come from that source person in <Video 1>.
<Video 1> is the original video being edited and defines the source temporal structure, performance, camera and lighting.
<Audio 1> is the enabled synchronized soundtrack of <Video 1>, reused 1:1 as the complete final audio track.

summary:
[video editing + reference generation + audio reuse] The target video is an edited version of <Video 1>. Replace the identity of each bound person using the assigned image: face, hairstyle, and body build within the original visible crop. Retain the original clothing and complete synchronized soundtrack.

retention\_analysis:
<Subject 1>: attribute\_transfer - transfer facial identity, hairstyle and visible body proportions from <Picture 1>; retain all source garments and accessories, pose, action, expression timing, gaze and interactions.
<Video 1>: partially\_preserved - change the bound facial identities, visible hairstyles and visible body build. Preserve source clothing, accessories, positions, performance, props, scene, camera, lighting, cuts and duration.
<Audio 1>: fully\_copy - reuse the entire synchronized source soundtrack without changing speech, voices, ambience, effects, music, pauses or timing.

detailed\_description:
[Shot 1] Start from the opening composition of <Video 1>. Track each bound source person by the locator above, through head turns, gestures and any occlusion. Apply the assigned facial identity and visible hairstyle consistently across the timeline, while following the original expression changes and speech articulation. Transfer body build and proportions only for body regions visible in both the source framing and the reference photograph. Do not infer unseen anatomy beneath reference clothing. Retain the source pose and root motion; fit the original garments to the changed build, preserving their identity, colour, material and styling. Keep source headwear and accessories; transfer hairstyle only in the regions that remain visible. Do not zoom out, uncrop, extend the body or invent unseen anatomy to demonstrate the reference. Keep person-to-person and hand-to-prop contacts coherent. Preserve all source cuts, camera angle, subject scale, framing, camera movement, lighting, background, objects and timing. Reuse <Audio 1> continuously as the complete final soundtrack. Preserve the original speech-to-person relationship, including any off-screen voices, without changing words, voices or timing.

overall\_soundscape:
Retain the speech, ambience and physical sounds already in <Audio 1>, at their original timing. Do not generate extra dialogue or assign off-screen speech to a visible person.

non\_diegetic\_music:
Retain any music already in <Audio 1>, as part of the complete copied track. Add no new score.

\endgroup

\subsection{Identity replacement with source-clothing retention -- Seedance 2.5}
\noindent Example A (Seedance 2.5).
\begingroup\small\raggedright
Reference and subject roles:
Subject 1 is the replacement for the lone foreground person in Video 1. Facial identity, hairstyle and visible body build come from the bound person in Image 1: the primary named person, centred in the photograph. Ignore all other people in that photograph. Clothing, headwear, accessories, pose and performance come from that source person in Video 1.
Video 1 is the original video being edited and defines the source temporal structure, performance, camera and lighting.
Video 1's original synchronized audio track is the enabled synchronized soundtrack of Video 1, reused 1:1 as the complete final audio track.

Target editing objective:
The target video is an edited version of Video 1. Replace the identity of each bound person using the assigned image: face, hairstyle, and body build within the original visible crop. Retain the original clothing and complete synchronized soundtrack.

Action, camera and audio progression:
Shot 1 Start from the opening composition of Video 1. Track each bound source person by the locator above, through head turns, gestures and any occlusion. Apply the assigned facial identity and visible hairstyle consistently across the timeline, while following the original expression changes and speech articulation. Transfer body build and proportions only for body regions visible in both the source framing and the reference photograph. Do not infer unseen anatomy beneath reference clothing. Retain the source pose and root motion; fit the original garments to the changed build, preserving their identity, colour, material and styling. Keep source headwear and accessories; transfer hairstyle only in the regions that remain visible. Do not zoom out, uncrop, extend the body or invent unseen anatomy to demonstrate the reference. Keep person-to-person and hand-to-prop contacts coherent. Preserve all source cuts, camera angle, subject scale, framing, camera movement, lighting, background, objects and timing. Reuse Video 1's original synchronized audio track continuously as the complete final soundtrack. Preserve the original speech-to-person relationship, including any off-screen voices, without changing words, voices or timing.

Ambient sound:
Retain the speech, ambience and physical sounds already in Video 1's original synchronized audio track, at their original timing. Do not generate extra dialogue or assign off-screen speech to a visible person.

Background music:
Retain any music already in Video 1's original synchronized audio track, as part of the complete copied track. Add no new score.

Preservation and modification requirements:
Subject 1: Transfer the specified attributes: transfer facial identity, hairstyle and visible body proportions from Image 1; retain all source garments and accessories, pose, action, expression timing, gaze and interactions.
Video 1: Preserve and modify within this scope: change the bound facial identities, visible hairstyles and visible body build. Preserve source clothing, accessories, positions, performance, props, scene, camera, lighting, cuts and duration.
Video 1's original synchronized audio track: Reuse in full: reuse the entire synchronized source soundtrack without changing speech, voices, ambience, effects, music, pauses or timing.

\endgroup

\subsection{Appearance generation -- binding excerpt}
\noindent Example B (MiniMax H3).
\begin{quote}\small subject\_definitions:
<Subject 1> is the person whose appearance is defined by <Picture 1>.
<Subject 2> is the person whose appearance is defined by <Picture 2>.\end{quote}

\subsection{Performance transfer -- binding excerpt}
\noindent Example C (MiniMax H3).
\begin{quote}\small subject\_definitions:
<Subject 1> is the person whose appearance is defined by <Picture 1>. His visual performance is supplied by the left standing performer in <Video 1>, whose own appearance identifies only the source role.
<Subject 2> is the person whose appearance is defined by <Picture 2>. His visual performance is supplied by the right seated performer in <Video 1>, whose own appearance identifies only the source role.
<Video 1> is the silent source of the whole-video framing, camera movement, shot structure, and temporal pacing.\end{quote}

\subsection{Timbre-reference generation -- binding excerpt}
\noindent Example D (MiniMax H3).
\begin{quote}\small summary:
[reference generation + audio reference] The target video shows <Subject 1> and <Subject 2> seated in <Subject 3>. <Subject 1> proposes assigning the task of capturing Chenxiang to Li Jing and his son, and <Subject 2> agrees. <Audio 1> and <Audio 2> provide the voice-timbre references for the two speakers.\end{quote}

In the original prompt, <Subject 3> denotes the interior setting---a decorative lattice screen with a circular floral pattern, dark wooden furniture, and warm lighting---rather than a third person.

\subsection{Appearance editing -- binding excerpt}
\noindent Example E (MiniMax H3).
\begin{quote}\small subject\_definitions:
<Subject 1> is the replacement person in the output, with facial identity from <Picture 1>. This person replaces the person initially on the left in <Video 1>, retaining that source person's clothing and performance.
<Subject 2> is the replacement person in the output, with facial identity from <Picture 2>. This person replaces the person initially on the right in <Video 1>, retaining that source person's clothing and performance.
<Video 1> is the source shot.
<Audio 1> is the complete synchronized soundtrack of <Video 1>.\end{quote}

\subsection{Joint reference generation -- binding excerpt}
\noindent Example F (MiniMax H3).
\begin{quote}\small subject\_definitions:
<Subject 1> is the person whose appearance is defined by <Picture 1>. His performance is provided by the standing older performer in <Video 1>, whose own appearance identifies only the source role.
<Subject 2> is the person whose appearance is defined by <Picture 2>. Her performance is provided by the seated younger performer in <Video 1>, whose own appearance identifies only the source role.
<Video 1> is the silent source of the whole-video framing, camera movement, shot structure, and temporal pacing.
<Audio 1> is the voice-timbre reference for <Subject 1> (S1); the reference recording's words are not copied.
<Audio 2> is the voice-timbre reference for <Subject 2> (S2); the reference recording's words are not copied.\end{quote}

\subsection{Appearance + voice editing -- binding excerpt}
\noindent Example G (MiniMax H3).
\begin{quote}\small summary:
[video editing + audio reference] The target video is an edited version of <Video 1>. It replaces the original characters with <Subject 1> and <Subject 2> while maintaining the original camera movement and scene composition. <Subject 2> (S2) delivers two lines of dialogue using the voice timbre from <Audio 2>, while <Subject 1> remains silent. <Audio 1> is available for <Subject 1> but is not used because that subject has no spoken line.\end{quote}

\subsection{Audio-driven lip editing -- binding excerpt}
\noindent Example H (MiniMax H3).
\begin{quote}\small subject\_definitions:
<Subject 1> is the person initially on the left in <Video 1>.
<Subject 2> is the person initially on the right in <Video 1>.
<Video 1> is the silent visual edit source.
<Audio 1> is the supplied final soundtrack and speech timing reference.\end{quote}

\subsection{Background editing -- binding excerpt}
\noindent Example I (MiniMax H3).
\begin{quote}\small subject\_definitions:
<Subject 1> is original foreground person 1, tracked consistently in <Video 1>, retained with the same identity, appearance, clothing and performance.
<Subject 2> is original foreground person 2, tracked consistently in <Video 1>, retained with the same identity, appearance, clothing and performance.
<Subject 3> is a replacement environment guided by <Picture 1>, whose representative features include a long corridor with pale walls, a red wall strip, rectangular doorways and a tiled floor. Use the photograph for the kind of place, characteristic structures, materials and furnishings. Its exact layout and individual details may be adapted to fit the source shot naturally.
<Video 1> is the original video being edited and defines the foreground performance, temporal structure, camera and lighting.
<Audio 1> is the enabled synchronized soundtrack of <Video 1>, reused 1:1 as the complete final audio track.\end{quote}

\end{document}

%% file: tables/condition_main.tex
\begin{table}[!t]
\centering
\caption{Task-condition measurements on metric-specific common cases. $C/A$ counts cases with complete in-scope records for all five models / applicable cases. Similarities retain their raw scales; pose is a 0--100 endpoint proxy. CER/WER are percentages. The final row equally averages positions over 16 task/subject groups using 503 rankable cases; rankability does not require every metric to be measured. Bold and underline mark the best and second-best displayed values, including ties.}
\label{tab:condition-main}
\fontsize{8.5}{10}\selectfont
\setlength{\tabcolsep}{3pt}
\begin{tabular*}{\linewidth}{@{\extracolsep{\fill}}llrrrrrr@{}}
\arrayrulecolor{jingdongred}\toprule\arrayrulecolor{black}
 \paperth{Task} & \paperth{Measurement} & \paperth{$C/A$} & \paperth{MiniMax H3} & \paperth{Seedance 2.0} & \paperth{Seedance 2.5} & \paperth{Kling 4.0} & \paperth{Wan 3.0} \\
\midrule
\paperth{Appearance} & Target identity $\uparrow$ & 82/107 & 0.686 & 0.751 & \underline{0.772} & 0.705 & \textbf{0.823} \\
 & Person DINO $\uparrow$ & 100/107 & \underline{0.836} & 0.834 & 0.826 & 0.833 & \textbf{0.887} \\
 & Garment DINO $\uparrow$ & 91/107 & \underline{0.749} & 0.747 & 0.719 & 0.720 & \textbf{0.803} \\
\midrule[0.2pt]
\paperth{Performance} & Target identity $\uparrow$ & 90/93 & 0.344 & \underline{0.782} & 0.775 & 0.742 & \textbf{0.825} \\
 & Endpoint pose $\uparrow$ & 49/72 & \textbf{54.300} & 39.000 & \underline{52.700} & 45.800 & 46.200 \\
\midrule[0.2pt]
\paperth{Voice} & Timbre cosine $\uparrow$ & 131/137 & \underline{0.601} & 0.551 & 0.505 & 0.315 & \textbf{0.622} \\
\paperth{reference} & CER (\%) $\downarrow$ & 137/137 & 4.620 & 4.070 & \underline{2.490} & \textbf{2.020} & 6.240 \\
\midrule[0.2pt]
\paperth{Person} & Target identity $\uparrow$ & 23/23 & 0.415 & 0.500 & 0.485 & \underline{0.553} & \textbf{0.696} \\
\paperth{replacement} & Source garment $\uparrow$ & 22/23 & 0.882 & \underline{0.912} & \textbf{0.916} & 0.718 & 0.798 \\
 & Appearance CLIP $\uparrow$ & 85/92 & 0.255 & 0.252 & 0.237 & \textbf{0.267} & \underline{0.257} \\
\midrule[0.2pt]
\paperth{Joint} & Target identity $\uparrow$ & 37/37 & 0.352 & 0.791 & \underline{0.816} & 0.766 & \textbf{0.836} \\
\paperth{reference} & Endpoint pose $\uparrow$ & 8/10 & \underline{46.100} & 29.200 & 26.700 & \textbf{46.200} & 32.400 \\
 & Timbre cosine $\uparrow$ & 33/37 & \textbf{0.585} & 0.541 & 0.490 & 0.275 & \underline{0.584} \\
 & CER (\%) $\downarrow$ & 37/37 & 6.070 & \textbf{3.970} & 5.420 & \underline{5.400} & 11.280 \\
\midrule[0.2pt]
\paperth{Appearance} & Appearance CLIP $\uparrow$ & 53/60 & 0.234 & 0.237 & 0.232 & \textbf{0.248} & \underline{0.241} \\
\paperth{+ voice} & Timbre cosine $\uparrow$ & 57/60 & \textbf{0.602} & 0.469 & 0.281 & 0.238 & \underline{0.531} \\
 & CER (\%) $\downarrow$ & 60/60 & 1.630 & \underline{1.490} & 1.650 & \textbf{1.000} & 2.400 \\
\midrule[0.2pt]
\paperth{Lip editing} & WER (\%) $\downarrow$ & 18/18 & 65.490 & 16.250 & \textbf{5.180} & 120.070 & \underline{13.510} \\
\midrule[0.2pt]
\paperth{Background} & Background difference $\uparrow$ & 35/39 & -0.113 & 0.008 & \underline{0.208} & \textbf{0.357} & 0.135 \\
 & Source identity $\uparrow$ & 31/39 & 0.892 & \underline{0.915} & \textbf{0.969} & 0.889 & 0.891 \\
 & Harmony $\uparrow$ & 36/39 & 0.643 & 0.688 & \textbf{0.708} & 0.684 & \underline{0.702} \\
\midrule
\multicolumn{2}{l}{\paperth{Overall mean rank $\downarrow$}} & 503/606 & 3.594 & \underline{3.000} & 3.250 & 3.531 & \textbf{1.625} \\
\bottomrule
\end{tabular*}
\end{table}

%% file: tables/condition_coverage.tex
\begin{table}[!t]\centering\footnotesize
\caption{Video-level coverage of task-condition measurements. A: applicable; F: numeric records without an explicit partial-scope flag; P: partial roles/turns or narrowed scope; U: unavailable; N: inapplicable; G: content-gated; Z: confirmed speech-absence failure. Each row satisfies $A=F+P+U+G+Z$ and $A+N=3{,}030$.}\label{tab:condition-coverage}
\begin{tabular}{@{}lrrrrrrr@{}}\arrayrulecolor{jingdongred}\toprule\arrayrulecolor{black}
 \paperth{Measurement} & \paperth{A} & \paperth{F} & \paperth{P} & \paperth{U} & \paperth{N} & \paperth{G} & \paperth{Z} \\\midrule
Target identity & 1300 & 1190 & 61 & 49 & 1730 & 0 & 0 \\
\paperrowsep
Person DINO & 1185 & 1154 & 10 & 21 & 1845 & 0 & 0 \\
\paperrowsep
Garment DINO & 1185 & 1102 & 45 & 38 & 1845 & 0 & 0 \\
\paperrowsep
Source ArcFace (transfer) & 765 & 640 & 0 & 125 & 2265 & 0 & 0 \\
\paperrowsep
Target--source ArcFace & 765 & 603 & 20 & 142 & 2265 & 0 & 0 \\
\paperrowsep
Source garment & 115 & 114 & 1 & 0 & 2915 & 0 & 0 \\
\paperrowsep
Appearance CLIP & 760 & 698 & 0 & 62 & 2270 & 0 & 0 \\
\paperrowsep
Face CLIP & 750 & 682 & 11 & 57 & 2280 & 0 & 0 \\
\paperrowsep
Hair CLIP & 655 & 588 & 7 & 60 & 2375 & 0 & 0 \\
\paperrowsep
Clothing CLIP & 655 & 602 & 1 & 52 & 2375 & 0 & 0 \\
\paperrowsep
Target background & 195 & 183 & 0 & 12 & 2835 & 0 & 0 \\
\paperrowsep
Source background & 195 & 183 & 0 & 12 & 2835 & 0 & 0 \\
\paperrowsep
Background difference & 195 & 183 & 0 & 12 & 2835 & 0 & 0 \\
\paperrowsep
Source identity & 195 & 167 & 0 & 28 & 2835 & 0 & 0 \\
\paperrowsep
Harmony & 195 & 189 & 0 & 6 & 2835 & 0 & 0 \\
\paperrowsep
First endpoint & 360 & 292 & 0 & 68 & 2670 & 0 & 0 \\
\paperrowsep
Last endpoint & 400 & 321 & 0 & 79 & 2630 & 0 & 0 \\
\paperrowsep
Endpoint pose & 410 & 303 & 48 & 59 & 2620 & 0 & 0 \\
\paperrowsep
CER (\%) & 1260 & 1260 & 0 & 0 & 1770 & 0 & 0 \\
\paperrowsep
WER (\%) & 1260 & 1200 & 0 & 60 & 1770 & 0 & 0 \\
\paperrowsep
Timbre cosine & 1170 & 955 & 186 & 0 & 1860 & 27 & 2 \\
\paperrowsep
Opening offset (ms) & 90 & 43 & 18 & 0 & 2940 & 29 & 0 \\
\paperrowsep
Sync $|$lag$|$ (ms) & 90 & 54 & 6 & 1 & 2940 & 29 & 0 \\
\paperrowsep
Recording waveform & 770 & 770 & 0 & 0 & 2260 & 0 & 0 \\
\paperrowsep
Recording correlation peak & 770 & 770 & 0 & 0 & 2260 & 0 & 0 \\
\paperrowsep
Recording log spectrum & 770 & 770 & 0 & 0 & 2260 & 0 & 0 \\
\paperrowsep
Recording duration error & 770 & 770 & 0 & 0 & 2260 & 0 & 0 \\
\paperrowsep
Background gen.-to-source & 465 & 457 & 0 & 8 & 2565 & 0 & 0 \\
\paperrowsep
Background source-to-gen. & 465 & 457 & 0 & 8 & 2565 & 0 & 0 \\
\paperrowsep
Generated background CLIP & 440 & 435 & 0 & 5 & 2590 & 0 & 0 \\
\paperrowsep
Source background CLIP & 440 & 435 & 0 & 5 & 2590 & 0 & 0 \\
\paperrowsep
Background CLIP gain & 440 & 432 & 0 & 8 & 2590 & 0 & 0 \\
\paperrowsep
Wrong-reference ArcFace & 60 & 60 & 0 & 0 & 2970 & 0 & 0 \\
\paperrowsep
Identity binding margin & 60 & 60 & 0 & 0 & 2970 & 0 & 0 \\
\paperrowsep
Visible-hair diagnostic & 20 & 13 & 2 & 5 & 3010 & 0 & 0 \\
\paperrowsep
Torso-geometry diagnostic & 20 & 0 & 0 & 20 & 3010 & 0 & 0 \\
\bottomrule\end{tabular}\end{table}

%% file: tables/lip_available.tex
\begin{table}[!t]\centering\footnotesize
\caption{Lip-editing measurements on system-specific available cases. Gate counts videos with WER $>50\%$ out of 18; $n$ counts complete in-scope numeric records, including flagged shorter anchors or sentence windows. Opening offsets are signed and sync uses absolute lag. Means use different case sets across systems.}\label{tab:lip-available}
\begin{tabular}{@{}lrrrrr@{}}\arrayrulecolor{jingdongred}\toprule\arrayrulecolor{black}
 \paperth{System} & \paperth{Gate} & \paperth{Opening $n$} & \paperth{Mean (ms)} & \paperth{Sync $n$} & \paperth{Mean $|$lag$|$ (ms)} \\\midrule
MiniMax H3 & 8 & 10 & 5.300 & 8 & 122.500 \\
\paperrowsep
Seedance 2.0 & 1 & 17 & -185.100 & 15 & 61.300 \\
\paperrowsep
Seedance 2.5 & 0 & 18 & -575.600 & 16 & 73.800 \\
\paperrowsep
Kling 4.0 & 18 & 0 & -- & 0 & -- \\
\paperrowsep
Wan 3.0 & 2 & 16 & 159.200 & 16 & 221.200 \\
\bottomrule\end{tabular}\end{table}

%% file: tables/condition_subject_zh.tex
\begin{table}[!t]\centering\footnotesize
\setlength{\tabcolsep}{3pt}
\caption{Task-condition measurements by subject setting: reference generation and person replacement. S/M denote single/multi; $C/A$ and units follow Table~\ref{tab:condition-main}. Bold and underline mark the best and second-best displayed values, including ties.}\label{tab:condition-subject}
\begin{tabular}{@{}llrrrrrr@{}}\arrayrulecolor{jingdongred}\toprule\arrayrulecolor{black}
 \paperth{Task/group} & \paperth{Measurement} & \paperth{$C/A$} & \paperth{\shortstack{MiniMax\\H3}} & \paperth{\shortstack{Seedance\\2.0}} & \paperth{\shortstack{Seedance\\2.5}} & \paperth{Kling 4.0} & \paperth{Wan 3.0} \\
\midrule
\paperth{Appearance (S)} & Target identity $\uparrow$ & 50/58 & 0.705 & 0.767 & \underline{0.796} & 0.743 & \textbf{0.838} \\
\paperrowsep
 & Person DINO $\uparrow$ & 55/58 & 0.819 & 0.813 & 0.811 & \underline{0.839} & \textbf{0.876} \\
\paperrowsep
 & Garment DINO $\uparrow$ & 52/58 & \underline{0.735} & 0.727 & 0.709 & 0.721 & \textbf{0.796} \\
\midrule[0.2pt]
\paperth{Appearance (M)} & Target identity $\uparrow$ & 32/49 & 0.655 & 0.725 & \underline{0.734} & 0.646 & \textbf{0.798} \\
\paperrowsep
 & Person DINO $\uparrow$ & 45/49 & 0.857 & \underline{0.860} & 0.844 & 0.824 & \textbf{0.900} \\
\paperrowsep
 & Garment DINO $\uparrow$ & 39/49 & 0.768 & \underline{0.774} & 0.733 & 0.719 & \textbf{0.813} \\
\midrule[0.2pt]
\paperth{Performance (S)} & Target identity $\uparrow$ & 39/41 & 0.382 & \underline{0.826} & 0.798 & 0.753 & \textbf{0.832} \\
\paperrowsep
 & Endpoint pose $\uparrow$ & 22/31 & \textbf{57.200} & 36.700 & \underline{56.100} & 47.800 & 55.700 \\
\midrule[0.2pt]
\paperth{Performance (M)} & Target identity $\uparrow$ & 51/52 & 0.315 & 0.748 & \underline{0.757} & 0.733 & \textbf{0.819} \\
\paperrowsep
 & Endpoint pose $\uparrow$ & 27/41 & \textbf{52.000} & 40.800 & \underline{49.900} & 44.100 & 38.400 \\
\midrule[0.2pt]
\paperth{Voice reference (S)} & Timbre cosine $\uparrow$ & 57/59 & \underline{0.612} & 0.581 & 0.507 & 0.340 & \textbf{0.623} \\
\paperrowsep
 & CER (\%) $\downarrow$ & 59/59 & 2.310 & 2.750 & \textbf{1.020} & \underline{1.250} & 3.620 \\
\midrule[0.2pt]
\paperth{Voice reference (M)} & Timbre cosine $\uparrow$ & 74/78 & \underline{0.593} & 0.528 & 0.502 & 0.295 & \textbf{0.622} \\
\paperrowsep
 & CER (\%) $\downarrow$ & 78/78 & 6.370 & 5.070 & \underline{3.600} & \textbf{2.600} & 8.230 \\
\midrule[0.2pt]
\paperth{Person replacement (S)} & Target identity $\uparrow$ & 11/11 & 0.510 & 0.491 & 0.543 & \underline{0.561} & \textbf{0.727} \\
\paperrowsep
 & Source garment $\uparrow$ & 11/11 & 0.885 & \underline{0.900} & \textbf{0.901} & 0.727 & 0.777 \\
\paperrowsep
 & Appearance CLIP $\uparrow$ & 50/52 & 0.247 & 0.247 & 0.231 & \textbf{0.261} & \underline{0.251} \\
\midrule[0.2pt]
\paperth{Person replacement (M)} & Target identity $\uparrow$ & 12/12 & 0.328 & 0.508 & 0.432 & \underline{0.545} & \textbf{0.667} \\
\paperrowsep
 & Source garment $\uparrow$ & 11/12 & 0.880 & \underline{0.924} & \textbf{0.932} & 0.709 & 0.819 \\
\paperrowsep
 & Appearance CLIP $\uparrow$ & 35/40 & \underline{0.266} & 0.258 & 0.246 & \textbf{0.276} & \underline{0.266} \\
\bottomrule\end{tabular}\end{table}
\FloatBarrier
\begin{table}[!t]\centering\footnotesize
\setlength{\tabcolsep}{3pt}
\caption{Task-condition measurements by subject setting: joint control, lip editing, and background replacement. Coverage, units, and best/second-best markings follow Table~\ref{tab:condition-subject}.}
\begin{tabular}{@{}llrrrrrr@{}}\arrayrulecolor{jingdongred}\toprule\arrayrulecolor{black}
 \paperth{Task/group} & \paperth{Measurement} & \paperth{$C/A$} & \paperth{\shortstack{MiniMax\\H3}} & \paperth{\shortstack{Seedance\\2.0}} & \paperth{\shortstack{Seedance\\2.5}} & \paperth{Kling 4.0} & \paperth{Wan 3.0} \\
\midrule
\paperth{Joint reference (S)} & Target identity $\uparrow$ & 19/19 & 0.375 & 0.746 & \underline{0.826} & 0.757 & \textbf{0.842} \\
\paperrowsep
 & Endpoint pose $\uparrow$ & 3/5 & \textbf{61.300} & 50.100 & 41.400 & 55.400 & \underline{60.600} \\
\paperrowsep
 & Timbre cosine $\uparrow$ & 16/19 & \textbf{0.595} & 0.550 & 0.485 & 0.293 & \underline{0.589} \\
\paperrowsep
 & CER (\%) $\downarrow$ & 19/19 & \underline{3.880} & 4.210 & \textbf{3.840} & 6.780 & 11.150 \\
\midrule[0.2pt]
\paperth{Joint reference (M)} & Target identity $\uparrow$ & 18/18 & 0.328 & \textbf{0.840} & 0.806 & 0.775 & \underline{0.830} \\
\paperrowsep
 & Endpoint pose $\uparrow$ & 5/5 & \underline{37.000} & 16.700 & 18.000 & \textbf{40.800} & 15.500 \\
\paperrowsep
 & Timbre cosine $\uparrow$ & 17/18 & \underline{0.576} & 0.533 & 0.494 & 0.259 & \textbf{0.578} \\
\paperrowsep
 & CER (\%) $\downarrow$ & 18/18 & 8.380 & \textbf{3.710} & 7.080 & \underline{3.940} & 11.410 \\
\midrule[0.2pt]
\paperth{Appearance + voice (S)} & Appearance CLIP $\uparrow$ & 28/28 & 0.212 & 0.211 & 0.214 & \textbf{0.221} & \underline{0.219} \\
\paperrowsep
 & Timbre cosine $\uparrow$ & 28/28 & \textbf{0.610} & 0.542 & 0.327 & 0.263 & \underline{0.583} \\
\paperrowsep
 & CER (\%) $\downarrow$ & 28/28 & 1.470 & \underline{0.850} & 0.880 & \textbf{0.620} & 3.070 \\
\midrule[0.2pt]
\paperth{Appearance + voice (M)} & Appearance CLIP $\uparrow$ & 25/32 & 0.258 & \underline{0.266} & 0.252 & \textbf{0.277} & 0.265 \\
\paperrowsep
 & Timbre cosine $\uparrow$ & 29/32 & \textbf{0.595} & 0.398 & 0.237 & 0.215 & \underline{0.481} \\
\paperrowsep
 & CER (\%) $\downarrow$ & 32/32 & \underline{1.780} & 2.050 & 2.320 & \textbf{1.330} & 1.810 \\
\midrule[0.2pt]
\paperth{Lip editing (S)} & WER (\%) $\downarrow$ & 10/10 & 85.86 & \underline{4.30} & \textbf{1.98} & 128.19 & 7.04 \\
\paperrowsep
 & Opening offset (ms) & 0/10 & -- & -- & -- & -- & -- \\
\paperrowsep
 & Sync $|$lag$|$ (ms) $\downarrow$ & 0/10 & -- & -- & -- & -- & -- \\
\midrule[0.2pt]
\paperth{Lip editing (M)} & WER (\%) $\downarrow$ & 8/8 & 40.02 & 31.20 & \textbf{9.19} & 109.91 & \underline{21.59} \\
\paperrowsep
 & Opening offset (ms) & 0/8 & -- & -- & -- & -- & -- \\
\paperrowsep
 & Sync $|$lag$|$ (ms) $\downarrow$ & 0/8 & -- & -- & -- & -- & -- \\
\midrule[0.2pt]
\paperth{Background (S)} & Background difference $\uparrow$ & 18/20 & -0.045 & -0.081 & \underline{0.225} & \textbf{0.347} & 0.133 \\
\paperrowsep
 & Source identity $\uparrow$ & 19/20 & 0.902 & 0.906 & \textbf{0.983} & 0.920 & \underline{0.922} \\
\paperrowsep
 & Harmony $\uparrow$ & 19/20 & 0.637 & 0.689 & \textbf{0.747} & 0.726 & \underline{0.729} \\
\midrule[0.2pt]
\paperth{Background (M)} & Background difference $\uparrow$ & 17/19 & -0.186 & 0.103 & \underline{0.190} & \textbf{0.368} & 0.136 \\
\paperrowsep
 & Source identity $\uparrow$ & 12/19 & 0.876 & \underline{0.928} & \textbf{0.948} & 0.841 & 0.842 \\
\paperrowsep
 & Harmony $\uparrow$ & 17/19 & 0.649 & \textbf{0.686} & 0.663 & 0.637 & \underline{0.672} \\
\bottomrule\end{tabular}\end{table}
\FloatBarrier

%% file: tables/condition_ranking_zh.tex
\subsection{Overall task-condition rank}
\label{sec:condition-ranking}
We aggregate sample-level comparisons using fixed task weights and comparison scales. The numerical inputs follow the evaluated conditions: timbre uses the mean raw cosine over the required measured roles, and timing uses the absolute opening offset or mean absolute in-scope synchronization lag. Post-gate diagnostic cosines and confirmed-absence condition zeros are excluded from numeric cosine comparisons. These states remain distinct in the records; their unknown numeric contributions retain bounds. Lip editing follows the dialogue gate: an ungated output precedes a gated output, and two gated outputs tie. Missing WER leaves that gate undecided and comparison proceeds with the available evidence.

For metric $k$, let $d_k\in\{-1,1\}$ indicate whether larger values are preferred, $s_k$ be its fixed difference scale, and $\epsilon_k$ its tolerance. For models $a,b$ on sample $i$, a comparable numeric pair contributes
\begin{equation}
q_{iabk}=\operatorname{sgn}(\delta_{iabk})\min\!\left(1,\max\!\left(0,\frac{|\delta_{iabk}|-\epsilon_k}{s_k-\epsilon_k}\right)\right),
\qquad\delta_{iabk}=d_k(x_{iak}-x_{ibk}).
\end{equation}
Let $\mathcal K_i$ contain the applicable terms, $\mathcal O_{iab}$ those with compatible numeric evidence and reference bindings, and $w_{ik}$ the task weights. The combined comparison is bounded by
\begin{equation}
[L_{iab},U_{iab}]=\frac{\sum_{k\in\mathcal O_{iab}}w_{ik}q_{iabk}\ \pm\sum_{k\in\mathcal K_i\setminus\mathcal O_{iab}}w_{ik}}{\sum_{k\in\mathcal K_i}w_{ik}}.
\end{equation}
An interval entirely above 0.05 gives a win; one entirely below $-0.05$ gives a loss; an interval contained in $[-0.05,0.05]$ gives a tie. Otherwise the comparison is unresolved. An absent hair/build calibration remains an unknown term. Text-defined replacement uses its fixed corroboration rule: the whole-person CLIP direction needs agreement from at least two of face, hair and clothing, with no opposite region; missing region directions are enumerated as $-1,0,1$. A result is assigned only when all feasible outcomes agree.

For a sample with ten resolved pairs, model $m$ receives $P_{im}=\sum_{n\ne m}(\mathbf 1[\mathrm{win}]+\tfrac12\mathbf 1[\mathrm{tie}])$ and rank
\begin{equation}
r_{im}=1+\sum_{n\ne m}\mathbf 1[P_{in}>P_{im}]+\tfrac12\sum_{n\ne m}\mathbf 1[P_{in}=P_{im}].
\end{equation}
Within each task/subject group $g$, we average $r_{im}$ over its common rankable samples. In ascending order, a tied tier extends while its largest and smallest averages differ by at most 0.20; its members receive the mean occupied position $\rho_{gm}$. The overall value is $R_m=|\mathcal G_+|^{-1}\sum_{g\in\mathcal G_+}\rho_{gm}$ over groups with rankable samples. This descriptive tolerance is used within groups, without a second tolerance grouping of the overall values. Rankable means all pair outcomes are determined, possibly from interval bounds; it does not require every metric to be measured.

\paragraph{Fixed parameters.}
The task weights are listed in Table~\ref{tab:condition-rank-weights}. The pairs $(s_k,\epsilon_k)$ are $(0.5,0.01)$ for CER/WER, $(1,0.05)$ for raw timbre cosine, $(100,5)$ for endpoint pose, $(400,40)$\,ms for both timing magnitudes, and $(1,0.02)$ for Harmony and recording affinities. Reference-image controls give $(0.974691,0.051435)$ for ArcFace, $(0.477492,0.036705)$ for person DINO, and $(0.540999,0.015271)$ for garment DINO; source-preservation terms use the corresponding scales. Background difference uses $(0.263377,0.032922)$. The CLIP scales for whole person, face, hair and clothing are 0.038893, 0.075485, 0.028700 and 0.067039, with tolerance equal to half the scale. Computation retains full precision. The task weights and some comparison scales are fixed design choices without independent calibration; image-derived scales are applied to video. The resulting rank summarizes relative evidence over the measured conditions.

\par\noindent\begin{minipage}{\linewidth}
\centering\footnotesize
\captionof{table}{Fixed task weights for within-case comparisons. Inapplicable terms are removed; unknown applicable terms retain their weights and uncertainty bounds.}
\label{tab:condition-rank-weights}
\begin{tabular*}{\linewidth}{@{\extracolsep{\fill}}lp{0.72\linewidth}@{}}
\arrayrulecolor{jingdongred}\toprule\arrayrulecolor{black}
\paperth{Task / variant} & \paperth{Terms and weights} \\
\midrule
Appearance generation & Target identity 0.550; person DINO 0.300; garment DINO 0.150 \\
\paperrowsep
Performance transfer & Pose 0.500; target identity 0.300; person DINO 0.120; garment DINO 0.080 \\
\paperrowsep
Voice reference & Raw timbre cosine 0.700; CER 0.300 \\
\paperrowsep
Text replacement & Whole-person CLIP, corroborated by at least two regions \\
\paperrowsep
Face replacement & Target identity 0.650; source garment 0.250; waveform 0.060; log-spectrum 0.040 \\
\paperrowsep
Identity/hair/build & Identity 0.450; source garment 0.200; hair 0.150; build 0.150; waveform 0.030; log-spectrum 0.020 \\
\paperrowsep
Joint reference & Identity 0.200; person DINO 0.100; garment DINO 0.050; pose 0.300; timbre 0.250; CER 0.100 \\
\paperrowsep
Appearance + voice & Whole-person CLIP 0.450; timbre 0.400; CER 0.150 \\
\paperrowsep
Lip editing & WER 0.500; opening magnitude 0.250; sync magnitude 0.250, after the dialogue gate \\
\paperrowsep
Background editing & Background difference 0.650; source identity 0.250; Harmony 0.100 \\
\bottomrule
\end{tabular*}
\end{minipage}\par\medskip

\par\noindent\begin{minipage}{\linewidth}
\centering\footnotesize
\captionof{table}{Task/subject group positions and their equally weighted overall mean (lower is better). S/M denote single/multi; $N$ gives cases with ranks for all five models / all cases in the group. Bold and underline mark the best and second-best positions, including ties.}
\label{tab:condition-rank-groups}
\begin{tabular*}{\linewidth}{@{\extracolsep{\fill}}llrrrrrr@{}}
\arrayrulecolor{jingdongred}\toprule\arrayrulecolor{black}
\paperth{Task} & \paperth{Subjects} & \paperth{$N$} & \paperth{\shortstack{MiniMax\\H3}} & \paperth{\shortstack{Seedance\\2.0}} & \paperth{\shortstack{Seedance\\2.5}} & \paperth{Kling 4.0} & \paperth{Wan 3.0} \\
\midrule
Appearance generation & S & 47/58 & \underline{3.500} & \underline{3.500} & \underline{3.500} & \underline{3.500} & \textbf{1.000} \\
\paperrowsep
Appearance generation & M & 28/49 & 4.000 & \underline{2.000} & 3.000 & 5.000 & \textbf{1.000} \\
\paperrowsep
Performance transfer & S & 31/41 & 5.000 & 4.000 & \underline{2.000} & 3.000 & \textbf{1.000} \\
\paperrowsep
Performance transfer & M & 38/52 & 5.000 & \underline{3.000} & \underline{3.000} & \underline{3.000} & \textbf{1.000} \\
\paperrowsep
Voice reference & S & 57/59 & \textbf{1.500} & \underline{3.000} & 4.000 & 5.000 & \textbf{1.500} \\
\paperrowsep
Voice reference & M & 74/78 & \textbf{1.500} & \underline{3.500} & \underline{3.500} & 5.000 & \textbf{1.500} \\
\paperrowsep
Person replacement & S & 59/63 & \underline{3.500} & \textbf{1.500} & 5.000 & \textbf{1.500} & \underline{3.500} \\
\paperrowsep
Person replacement & M & 43/52 & \underline{4.500} & \textbf{2.000} & \underline{4.500} & \textbf{2.000} & \textbf{2.000} \\
\paperrowsep
Joint reference & S & 14/19 & 4.500 & \underline{2.500} & \underline{2.500} & 4.500 & \textbf{1.000} \\
\paperrowsep
Joint reference & M & 16/18 & 5.000 & \textbf{1.500} & \underline{3.000} & 4.000 & \textbf{1.500} \\
\paperrowsep
Appearance + voice & S & 28/28 & \textbf{1.500} & \underline{3.000} & 4.500 & 4.500 & \textbf{1.500} \\
\paperrowsep
Appearance + voice & M & 22/32 & \textbf{1.500} & \underline{3.500} & 5.000 & \underline{3.500} & \textbf{1.500} \\
\paperrowsep
Lip editing & S & 10/10 & 4.000 & \underline{2.500} & \underline{2.500} & 5.000 & \textbf{1.000} \\
\paperrowsep
Lip editing & M & 5/8 & \underline{3.000} & 4.000 & \textbf{1.500} & 5.000 & \textbf{1.500} \\
\paperrowsep
Background editing & S & 18/20 & 4.500 & 4.500 & \underline{2.000} & \textbf{1.000} & 3.000 \\
\paperrowsep
Background editing & M & 13/19 & 5.000 & 4.000 & \underline{2.500} & \textbf{1.000} & \underline{2.500} \\
\midrule
\multicolumn{2}{l}{\paperth{Overall mean rank $\downarrow$}} & 503/606 & 3.594 & \underline{3.000} & 3.250 & 3.531 & \textbf{1.625} \\
\bottomrule
\end{tabular*}
\end{minipage}\par\medskip
\FloatBarrier

%% file: tables/qwen_checklist_groups.tex
\begin{table}[!t]
\centering
\caption{Checklist group positions and common-case coverage. Positions follow within-case task-score ranking and the 0.20 group tie tolerance; lower is better. The two lip-editing groups have no complete task comparison across all five models.}
\label{tab:qwen-checklist-groups}
\fontsize{8.5}{10}\selectfont
\setlength{\tabcolsep}{3pt}
\begin{tabular*}{\linewidth}{@{\extracolsep{\fill}}llrrrrrr@{}}
\arrayrulecolor{jingdongred}\toprule\arrayrulecolor{black}
Task & Subjects & $C/N$ & MiniMax H3 & Seedance 2.0 & Seedance 2.5 & Kling 4.0 & Wan 3.0 \\
\midrule
Appearance & Single & 58/58 & 4.500 & \underline{2.500} & 4.500 & \underline{2.500} & \textbf{1.000} \\
\paperrowsep
Appearance & Multi & 49/49 & \underline{4.000} & \textbf{2.000} & 5.000 & \textbf{2.000} & \textbf{2.000} \\
\paperrowsep
Performance & Single & 41/41 & 5.000 & \underline{4.000} & \textbf{2.000} & \textbf{2.000} & \textbf{2.000} \\
\paperrowsep
Performance & Multi & 52/52 & 5.000 & \textbf{2.000} & \textbf{2.000} & \textbf{2.000} & \underline{4.000} \\
\paperrowsep
Voice reference & Single & 59/59 & \underline{3.500} & \underline{3.500} & \textbf{1.500} & \textbf{1.500} & 5.000 \\
\paperrowsep
Voice reference & Multi & 75/78 & \underline{3.500} & \underline{3.500} & \textbf{1.500} & \textbf{1.500} & 5.000 \\
\paperrowsep
Person replacement & Single & 63/63 & \underline{4.000} & \textbf{1.500} & \underline{4.000} & \textbf{1.500} & \underline{4.000} \\
\paperrowsep
Person replacement & Multi & 52/52 & \underline{3.000} & \underline{3.000} & 5.000 & \textbf{1.000} & \underline{3.000} \\
\paperrowsep
Joint reference & Single & 17/19 & 5.000 & \underline{3.500} & \textbf{1.500} & \textbf{1.500} & \underline{3.500} \\
\paperrowsep
Joint reference & Multi & 17/18 & 5.000 & \textbf{2.000} & \textbf{2.000} & \textbf{2.000} & \underline{4.000} \\
\paperrowsep
Appearance + voice & Single & 27/28 & \underline{4.500} & \textbf{2.000} & \underline{4.500} & \textbf{2.000} & \textbf{2.000} \\
\paperrowsep
Appearance + voice & Multi & 29/32 & \textbf{1.000} & \underline{3.500} & \underline{3.500} & \underline{3.500} & \underline{3.500} \\
\paperrowsep
Lip editing & Single & 0/10 & -- & -- & -- & -- & -- \\
\paperrowsep
Lip editing & Multi & 0/8 & -- & -- & -- & -- & -- \\
\paperrowsep
Background & Single & 20/20 & 5.000 & \underline{4.000} & \textbf{2.000} & \textbf{2.000} & \textbf{2.000} \\
\paperrowsep
Background & Multi & 19/19 & 5.000 & 4.000 & \textbf{1.500} & \textbf{1.500} & \underline{3.000} \\
\bottomrule
\end{tabular*}
\end{table}

%% file: tables/qwen_flashnext_groups.tex
\begin{table}[!t]
\centering
\caption{Qwen3.8-Flash-Next-FP8 checklist group positions and common-case coverage. Positions follow within-case task-score ranking and the 0.20 group tie tolerance; lower is better. The two lip-editing groups have no complete task comparison across all five models.}
\label{tab:qwen-flashnext-groups}
\fontsize{8.5}{10}\selectfont
\setlength{\tabcolsep}{3pt}
\begin{tabular*}{\linewidth}{@{\extracolsep{\fill}}llrrrrrr@{}}
\arrayrulecolor{jingdongred}\toprule\arrayrulecolor{black}
Task & Subjects & $C/N$ & MiniMax H3 & Seedance 2.0 & Seedance 2.5 & Kling 4.0 & Wan 3.0 \\
\midrule
Appearance & Single & 57/58 & \textbf{3.000} & \textbf{3.000} & \textbf{3.000} & \textbf{3.000} & \textbf{3.000} \\
\paperrowsep
Appearance & Multi & 48/49 & \underline{4.500} & \textbf{2.000} & \underline{4.500} & \textbf{2.000} & \textbf{2.000} \\
\paperrowsep
Performance & Single & 41/41 & 5.000 & \underline{2.500} & \underline{2.500} & \textbf{1.000} & 4.000 \\
\paperrowsep
Performance & Multi & 52/52 & \underline{5.000} & \textbf{2.500} & \textbf{2.500} & \textbf{2.500} & \textbf{2.500} \\
\paperrowsep
Voice reference & Single & 57/59 & \textbf{3.000} & \textbf{3.000} & \textbf{3.000} & \textbf{3.000} & \textbf{3.000} \\
\paperrowsep
Voice reference & Multi & 45/78 & \textbf{1.500} & \underline{4.000} & \textbf{1.500} & \underline{4.000} & \underline{4.000} \\
\paperrowsep
Person replacement & Single & 61/63 & 4.500 & \textbf{1.500} & 4.500 & \textbf{1.500} & \underline{3.000} \\
\paperrowsep
Person replacement & Multi & 50/52 & \underline{5.000} & \textbf{2.500} & \textbf{2.500} & \textbf{2.500} & \textbf{2.500} \\
\paperrowsep
Joint reference & Single & 1/19 & \textbf{2.500} & \textbf{2.500} & \underline{5.000} & \textbf{2.500} & \textbf{2.500} \\
\paperrowsep
Joint reference & Multi & 1/18 & \underline{5.000} & \textbf{2.500} & \textbf{2.500} & \textbf{2.500} & \textbf{2.500} \\
\paperrowsep
Appearance + voice & Single & 1/28 & 5.000 & \textbf{2.000} & \underline{4.000} & \textbf{2.000} & \textbf{2.000} \\
\paperrowsep
Appearance + voice & Multi & 13/32 & \textbf{2.500} & \textbf{2.500} & \underline{5.000} & \textbf{2.500} & \textbf{2.500} \\
\paperrowsep
Lip editing & Single & 0/10 & -- & -- & -- & -- & -- \\
\paperrowsep
Lip editing & Multi & 0/8 & -- & -- & -- & -- & -- \\
\paperrowsep
Background & Single & 18/20 & 5.000 & 4.000 & \textbf{1.000} & \underline{2.500} & \underline{2.500} \\
\paperrowsep
Background & Multi & 16/19 & 5.000 & 4.000 & \textbf{1.500} & \textbf{1.500} & \underline{3.000} \\
\bottomrule
\end{tabular*}
\end{table}

%% file: tables/preservation_variants_zh.tex
\begin{table}[!t]\centering\footnotesize\setlength{\tabcolsep}{3pt}
\caption{Person-replacement measurements by editing variant. $C/A$ gives common/applicable cases. Identity margin compares assigned and competing references; recording affinities cover the compared audio interval. Higher is better within each row; bold and underline mark the best and second-best displayed values, including ties.}\label{tab:preservation-diagnostics}
\begin{tabular}{@{}llrrrrrr@{}}\arrayrulecolor{jingdongred}\toprule\arrayrulecolor{black}
 \paperth{Variant} & \paperth{Measurement} & \paperth{$C/A$} & \paperth{\shortstack{MiniMax\\H3}} & \paperth{\shortstack{Seedance\\2.0}} & \paperth{\shortstack{Seedance\\2.5}} & \paperth{Kling 4.0} & \paperth{Wan 3.0} \\\midrule
\paperth{Text appearance} & Face CLIP & 83/90 & 0.242 & 0.242 & 0.230 & \textbf{0.253} & \underline{0.249} \\
\paperrowsep
 & Hair CLIP & 82/89 & \underline{0.222} & \underline{0.222} & 0.221 & \textbf{0.224} & 0.217 \\
\paperrowsep
 & Clothing CLIP & 84/89 & \underline{0.254} & 0.252 & 0.243 & \textbf{0.265} & 0.248 \\
\midrule[0.2pt]
\paperth{Face replacement} & Identity margin & 10/10 & 0.193 & 0.449 & 0.354 & \underline{0.455} & \textbf{0.574} \\
\paperrowsep
 & Source garment & 19/19 & 0.883 & \underline{0.918} & \textbf{0.931} & 0.690 & 0.789 \\
\paperrowsep
 & Waveform cosine & 19/19 & 0.502 & \textbf{0.741} & 0.495 & 0.232 & \underline{0.708} \\
\paperrowsep
 & Log-spectral cosine & 19/19 & 0.725 & \underline{0.921} & \textbf{0.929} & 0.777 & 0.903 \\
\midrule[0.2pt]
\paperth{Identity/hair/build} & Identity margin & 2/2 & 0.236 & 0.372 & 0.160 & \underline{0.386} & \textbf{0.587} \\
\paperrowsep
 & Source garment & 3/4 & \underline{0.880} & 0.874 & 0.821 & \textbf{0.897} & 0.853 \\
\paperrowsep
 & Waveform cosine & 4/4 & 0.420 & 0.693 & \underline{0.703} & 0.386 & \textbf{0.923} \\
\paperrowsep
 & Log-spectral cosine & 4/4 & 0.678 & 0.821 & \underline{0.946} & 0.906 & \textbf{0.958} \\
\bottomrule\end{tabular}\end{table}